%% file: list-arxiv.tex
\documentclass[oneside,11pt]{article} 

\renewenvironment{abstract}
  {{\centering\large\bfseries Abstract\par}\vspace{0.7ex}%
    \bgroup
       \leftskip 20pt\rightskip 20pt\small\noindent\ignorespaces}%
  {\par\egroup\vskip 0.25ex}

{\par\egroup\vskip 0.25ex}

\usepackage[utf8]{inputenc} 
\usepackage[T1]{fontenc}    
\usepackage{hyperref}       
\usepackage{url}            
\usepackage{booktabs}       
\usepackage{amsfonts}       
\usepackage{nicefrac}       
\usepackage{microtype}      
\usepackage{xcolor}         
\usepackage{dsfont}
\usepackage{xspace}

\usepackage{algorithm,algorithmic}
\usepackage{amssymb,amsmath,wrapfig,amsthm} 
\usepackage[autostyle]{csquotes}
\usepackage{graphicx}
\numberwithin{equation}{section}

\newcommand{\zeronorm}[1]{\left\lVert #1 \right\rVert_{0}}
\newcommand{\twonorm}[1]{\left\lVert #1 \right\rVert_{2}}
\newcommand{\onenorm}[1]{\left\lVert #1 \right\rVert_{1}}

\newcommand{\nuclearnorm}[1]{\left\lVert #1 \right\rVert_{*}}
\newcommand{\fronorm}[1]{\left\lVert #1 \right\rVert_{F}}

\newcommand{\infnorm}[1]{\left\lVert #1 \right\rVert_{\infty}}

\newtheorem{theorem}{Theorem}
\newtheorem{lemma}[theorem]{Lemma}
\newtheorem{proposition}[theorem]{Proposition}

\theoremstyle{definition}
\newtheorem{remark}[theorem]{Remark}
\newtheorem{definition}[theorem]{Definition}

\newtheorem{fact}[theorem]{Fact}
\newtheorem{claim}[theorem]{Claim}

\newcommand{\lFloor}{\left\lfloor}
\newcommand{\rFloor}{\right\rfloor}

\newcommand{\abs}[1]{\left\lvert #1 \right\rvert}
\DeclareMathOperator{\Var}{Var}

\newcommand{\trace}[1]{\mathrm{Tr}\(#1\)}

\newcommand{\hermite}{\mathrm{He}}
\newcommand{\homogeneous}{\mathrm{Hom}}
\newcommand{\hard}{\mathrm{trim}}
\DeclareMathOperator{\E}{\mathbb{E}}
\newcommand{\R}{\mathbb{R}}
\newcommand{\I}{\mathbb{I}}

\newcommand{\N}{{N}}

\newcommand{\assign}{\leftarrow}

\newcommand{\poly}{\mathrm{poly}}
\newcommand{\polylog}{\mathrm{polylog}}

\newcommand{\calL}{\mathcal{L}}

\newcommand{\ba}{\boldsymbol{a}}

\renewcommand{\(}{\left(}
\renewcommand{\)}{\right)}
\renewcommand{\[}{\left[}
\renewcommand{\]}{\right]}

\newcommand{\Sgood}{S_{{G}}}

\newcommand{\sparseP}{\mathbb{P}(\R^d, l, \kappa, \psi)}

\newcommand{\cluster}{\textsc{Cluster}\xspace}
\newcommand{\aefilter}{\textsc{Attribute-efficient-Multifilter}\xspace}
\newcommand{\bmf}{\textsc{BasicMF}\xspace}
\newcommand{\harmf}{\textsc{HarmonicMF}\xspace}
\newcommand{\mmf}{\textsc{MultilinearMF}\xspace}
\newcommand{\deghom}{\textsc{Degree2Homogeneous}\xspace}
\newcommand{\listred}{\textsc{ListReduction}\xspace}

\newcommand{\ANS}{\mathrm{ANS}}
\newcommand{\NO}{\mathrm{NO}}

\newcommand{\TBD}{\mathrm{TBD}}

\newcommand{\lambdasparse}{\lambda^{*}_{\mathrm{sparse}}}

\newcommand{\tensor}{\mathrm{tensor}}

\usepackage{enumitem}

\title{List-Decodable Sparse Mean Estimation}
\author{
{Shiwei Zeng}\\
\texttt{szeng4@stevens.edu}\\
Stevens Institute of Technology
\and
{Jie Shen}\\
\texttt{jie.shen@stevens.edu}\\
Stevens Institute of Technology
}
\date{\today}
\begin{document}
\maketitle

\input{intro.tex}

\input{related.tex}
\input{pre.tex}

\input{algorithm.tex}

\input{conclusion.tex}


\clearpage

\bibliography{../../jshen_ref,../../szeng_ref}
\bibliographystyle{alpha}

\newpage
\appendix

\input{appendix.tex}

\end{document}

%% file: intro.tex
\begin{abstract}
Robust mean estimation is one of the most important problems in statistics: given a set of samples in $\mathbb{R}^d$ where an $\alpha$ fraction are drawn from some distribution $D$ and the rest are adversarially corrupted, we aim to estimate the mean of $D$. A surge of recent research interest has been focusing on the list-decodable setting where $\alpha \in (0, \frac12]$, and the goal is to output a finite number of estimates among which at least one approximates the target mean. In this paper, we consider that the underlying distribution $D$ is Gaussian with $k$-sparse mean. Our main contribution is the first polynomial-time algorithm that enjoys sample complexity $O\big(\mathrm{poly}(k, \log d)\big)$, i.e. poly-logarithmic in the dimension. One of our core algorithmic ingredients is using low-degree {\em sparse polynomials} to filter outliers, which may find more applications.
\end{abstract}

%
%
%
%
%
%
%
%
%
%
%
%
%

\section{Introduction}

Mean estimation is arguably a fundamental inference task in statistics and machine learning. Given a set of samples $\{x_1, \dots, x_n\} \subset \mathbb{R}^d$ where an $\alpha$ fraction are drawn from some well-behaved (e.g. Gaussian) distribution $D$ and the rest are adversarially corrupted, the goal is to estimate the mean of $D$. In the noiseless case where $\alpha = 1$, the problem can be easily solved in view of the concentration of measure phenomenon \cite{ledoux1991probability}. However, this is rarely the case  as modern data sets are often contaminated by random noise or even by adversarial corruptions. Thus, a great deal of recent efforts are focused on efficiently and robustly estimating the target mean in the presence of outliers.

Generally speaking, there is a phase transition between $\alpha > 1/2$ and $0 < \alpha \leq 1/2$, and solving either problem in a computationally efficient manner is highly nontrivial. The problem that most of the samples are uncorrupted, i.e. $\alpha > 1/2$, has a very long history dating back to the 1960s \cite{tukey1960survey,huber1964robust}, yet only until recently have computationally efficient algorithms been established~\cite{diakonikolas2016robust,lai2016agnostic}. The other yet more challenging regime concerns that an overwhelming fraction of the samples are corrupted, i.e. $\alpha \leq 1/2$, which even renders estimation impossible. This motivates a line of research on {\em list-decodable} mean estimation~\cite{charika2017learning}, where in place of outputting one single estimate, the algorithm is allowed to generate a finite list of candidates and is considered to be successful if there exists at least one candidate in the list that is sufficiently close to the target mean. 

In this work, we investigate the problem of list-decodable mean estimation, for which there have been a plethora of elegant results established in recent years. From a high level, most of them concern  error guarantees and running time. For example, \cite{charika2017learning} proposed the first tractable algorithm based on semidefinite programming, which runs in polynomial time and achieves optimal error rate for variance-bounded distributions. \cite{dia2018list} developed a multi-filtering scheme and showed that the error rate can be improved by using high degree polynomials if the underlying distribution is Gaussian. The more recent works~\cite{cmy2020list,dia2021list} further addressed the computational efficiency of this task and achieved almost linear running time in certain regimes.

Although all of these algorithms exhibit near-optimal guarantees on either error rate or computational complexity, it turns out that less is explored to improve another yet important metric: the sample complexity. In particular, the sample complexity of all these algorithms is $O(\poly(d))$, hence they quickly break down for data-demanding applications such as healthcare where the number of available samples is typically orders of magnitude less than the dimension $d$  \cite{wainwright2019high}. Therefore, a pressing question that needs to be addressed in such a high-dimensional regime is the following:

\begin{quote}
	Does there exist a provably robust algorithm for list-decodable mean estimation that runs in polynomial time and enjoys a sample complexity bound of $O(\polylog(d))$?
\end{quote}


In this paper, we answer the question in the affirmative by showing that when the target mean  is $k$-sparse, i.e. it has at most $k$ non-zero elements, it is {\em attribute-efficiently} list-decodable.

\begin{theorem}[Main result]\label{thm:main_meanEst}
Given parameter $\alpha\in(0,\frac12]$,  failure probability $\tau \in (0, 1)$, a natural number $\ell \geq 1$, and a set $T$ of  $\Omega\big(\frac{  \ell^7 \cdot k^{14\ell}}{\alpha^7} \cdot \log^{8\ell}(\frac{ \ell d}{\alpha \tau})\big)$ 
samples in $\R^d$, of which at least a $(2\alpha)$-fraction are independent draws from the Gaussian distribution $\N(\mu,\I_d)$ where $\zeronorm{\mu}\leq k$, there exists an algorithm that runs in time $\poly\(\abs{T}, d^{\ell}, \frac{1}{\alpha}\)$, uses polynomials of degree at most $2\ell$, and returns a list of $O(1/\alpha)$ number of $k$-sparse vectors such that with probability  $1-\tau$, the list contains at least one $\hat{\mu}\in\R^d$ with $\twonorm{\hat{\mu}-\mu} = {\tilde{O}\big(\alpha^{-\frac{1}{2\ell}} \cdot \sqrt{\ell}(\ell+\log\frac{1}{\alpha})\big) }$,  where $\tilde{O}(\cdot)$ hides poly-logarithmic factors.
\end{theorem}




\begin{remark}
The key message of the theorem is that when the true mean is $k$-sparse, it is possible to efficiently approximate it with $O(\polylog (d))$ samples. This is in stark contrast to existing list-decodable results \cite{charika2017learning,dia2018list,cmy2020list,dia2020list,dia2021list} where the sample complexity is $O(\poly(d))$. The only attribute-efficient robust mean estimators are \cite{balakrishnan2017computation,dia2019outlier,cheng2021outlier}, but their results  hold only for the mild corruption regime where $\alpha > 1/2$.
\end{remark}

\begin{remark}
Our algorithm and analysis hold for any degree $\ell \geq 1$. When $\ell=1$, the sample complexity reads as $\tilde{O}(\alpha^{-7} k^{14} \log^8 d)$ and the algorithm achieves error $\tilde{O}({\alpha}^{-\frac12})$. 
As opposed to an $O(1-\alpha)$ error rate obtained for $\alpha > 1/2$, the (non-vanishing) error rate  $\tilde{O}({\alpha}^{-\frac12})$ is typically what one can expect for list-decodable mean estimation under bounded second order moment condition, in light of the lower bounds in \cite{dia2018list}. 
When leveraging degree-$2\ell$ polynomials into algorithmic design, we obtain the improved ${\tilde{O}\big(\alpha^{-\frac{1}{2\ell}}\sqrt{\ell}(\ell+\log\frac{1}{\alpha}) \big) }$ error guarantee. Specially, when taking $\ell=\Theta\big(\log\frac{1}{\alpha}\big)$, our algorithm achieves error rate of $\tilde{O}\big(\log^{\frac32}(\frac{1}{\alpha})\big)$ in quasi-polynomial time. This is very close to the minimax error rate of $\Theta(\log^{\frac12}(\frac{1}{\alpha}))$ established in \cite{dia2018list}.
\end{remark}
\begin{remark}
If we further increase the sample size with an $\ell^\ell$ multiplicative factor with $\ell=\Theta(\log\frac{1}{\alpha})$, our algorithm will achieve an $\tilde{O}\big(\log^{\frac12}(\frac{1}{\alpha})\big)$ error guarantee, which matches the minimax lower bound. The proof follows the same pipeline and we leave it to interested readers.
\end{remark}

\subsection{Overview of Our Techniques}\label{subsec:overview}

Our main algorithm is inspired by the multifiltering framework of \cite{dia2018list}, where the primary idea is to  construct a sequence of polynomials to test the concentration of the samples to Gaussian so that the algorithm either certifies that the sample set behaves like Gaussian, or sanitizes it by removing a sufficient amount of outliers. 
Our key technical contribution lies into a new design of {\em sparse polynomials}, and new filtering rules tailored to the sparse polynomials.

\vspace{0.1in}\noindent
{\bfseries Sparse polynomials and sparsity-induced filters.} \  
To ensure that our algorithm is attribute-efficient, we will only control the maximum eigenvalue of the sample covariance matrix on sparse directions. Since such computation is NP-hard in general, we first consider a sufficient condition which tests the maximum Frobenuis norm under a cardinality constraint, similar to the idea of \cite{dia2019outlier}. If such Frobenuis norm is small, it implies a small restricted eigenvalue and hence the sample mean is returned. Otherwise, we construct sparse polynomials in the sense that they can be represented by a set of $O(\ell^2 k^{4\ell})$ basis polynomials and $O(\ell k^{2\ell})$ coordinates of the samples (see Definition~\ref{def:sparse_polynomial}), and measure the concentration of these sparse polynomials to the Gaussian. Now as the underlying polynomials are sparse, we also design new sparsity-induced filters to certify the sample set, as otherwise a large amount of clean samples will be removed. See Algorithm~\ref{alg:mainSub-highdegree} and Algorithm~\ref{alg:basic}.

\vspace{0.1in}\noindent
{\bfseries Clustering by $L_\infty$-norm.} \ 
Technically, the success of our attribute-efficient multifiltering approach hinges on a condition that all the samples lie within a small $L_{\infty}$-norm ball. It is not hard to see that all the Gaussian samples satisfy such condition, and we show that there is a simple scheme which can simultaneously prune and cluster the given samples into $O(1/\alpha)$ groups, such that the retained samples are close under the $L_{\infty}$-norm and at least one group contains most of the Gaussian samples. We note that the use of the $L_{\infty}$-norm as our metric ensures attribute efficiency of this step. An immediate implication of  this clustering step is that the polynomials of Gaussian samples will be close enough, which facilitates the analysis of the performance of our filters. See Section~\ref{sec:cluster}.

%% file: related.tex
\subsection{Related Works}\label{sec:related}

Breaking the barrier of the typical $O(\poly(d))$  sample complexity bound is one of the central problems across many fields of science and engineering. Motivated by real-world applications, a property termed {sparsity} is often assumed for this end, meaning that only $k$ out of the $d$ number of attributes contribute to the underlying inference problem. In this way, an improved bound of $O(\poly(k, \log d))$ can be obtained in many inference paradigms such as linear regression~\cite{chen1998atomic,tibshirani1996regression,candes2005decoding,donoho2006compressed,shen2017iteration,shen2017partial,shen2018tight,wang2018provable}, learning of threshold functions~\cite{littlestone1987learning,blum1995learning,servedio2012attribute,plan2013robust,awasthi2016learning,zhang2020efficient,shen2020one,shen2021attribute,shen2021sample}, principal component analysis~\cite{ma2013sparse,dia2019outlier}, and mean estimation~\cite{balakrishnan2017computation,dia2019outlier,cheng2021outlier}. Unfortunately, the success of all these attribute-efficient algorithms hinges on the presumption that the majority of the data are uncorrupted. 

\vspace{0.1in}\noindent
{\bfseries Learning with mild corruption ($\alpha > 1/2$).}
Learning in the presence of noise has been extensively studied in a broad context. In supervised learning where a sample consists of an instance (i.e. feature vector) and a label, lots of research efforts were dedicated to robust algorithms under label noise~\cite{angluin1987learning,sloan1988types,massart2006risk}. Recent years have witnessed significant progress towards optimal algorithms in the presence of label noise, see for example, \cite{kalai2005agnostic,awasthi2017power,diakonikolas2020learning,zhang2020efficient,diakonikolas2020polynomial,zeng2022crowd} and the  references therein. The regime that both instances and labels are corrupted turns out to be significantly more challenging. The problem of learning halfspaces under such setting was put forward in the 1980s \cite{valiant1985learning,kearns1988learning}, yet only until recently have efficient algorithms been established with near-optimal noise tolerance \cite{awasthi2017power,diakonikolas2018learning,shen2021sample,shen2021attribute}. In addition, \cite{bhatia2015robust,klivans2018efficient,liu2020high} studied robust linear regression and \cite{balakrishnan2017computation} presented a set of interesting results under various statistical models. More in line with this work is the problem of robust mean estimation, see the breakthrough works of \cite{diakonikolas2016robust,lai2016agnostic} and many follow-up works~\cite{dia2017being,balakrishnan2017computation,diakonikolas2017statistical,steinhardt2018resilience,kothari2018robust,dia2019outlier,hopkins2020robust,cheng2021outlier}.

\vspace{0.1in}\noindent
{\bfseries Learning with overwhelming corruption ($\alpha \leq 1/2$).}
The agnostic label noise of \cite{haussler1992decision,kearns1992efficient} seems the earliest model that allows the adversary to arbitrarily corrupt any fraction of the data (say $70\%$), though it can only corrupt labels. Following \cite{charika2017learning}, a considerate number of of recent works have studied the scenario that both instances and labels are grossly corrupted, and the goal is to output a finite list of candidate parameters among which at least one is a good approximation to the target. This includes list-decodable  learning of mixture models~\cite{dia2018list,dia2021cluster}, regression~\cite{karmalkar2019list,raghavendra2020list}, and subspace recovery~\cite{raghavendra2020subspace,bakshi2021list}.  Interestingly, there are some works studying the problem under  crowdsourcing models, where the samples are collected from crowd workers and most of them behave adversarially~\cite{steinhardt2016avoid,awasthi2017efficient,meister2018data,zeng2021semi}. 

It is worth noting that \cite{diaLDsparse2022} concurrently and independently developed a polynomial-time algorithm to solve the same problem, with an interesting difference-of-pairs metric to filter outliers. 

\subsection{Roadmap}

We  collect useful notations, definitions, and some preliminary results in Section~\ref{sec:pre}. Our main algorithms are described in Section~\ref{sec:algorithm} along with performance guarantees. We conclude the work in Section~\ref{sec:conclusion}, and defer all proof details to the appendix.

%% file: pre.tex

\section{Preliminaries}\label{sec:pre}

{\bfseries Vector, matrix, and tensor.} \  For a $d$-dimensional vector $v = (v_1, \dots, v_d)$, denote by $\twonorm{v}$ its $L_2$-norm, $\onenorm{v}$ its $L_1$-norm, $\zeronorm{v}$ its $L_0$-``norm'' that counts the number of non-zeros, and $\infnorm{v}$ its infinity norm. The hard thresholding operator $\hard_k:\R^d\rightarrow\R^d$ keeps the $k$ largest elements (in magnitude) of a vector and sets the remaining to zero. Let $[d]:=\{1,2,\dots,d\}$ for some natural number $d> 0$. For an index set $\Omega\subseteq[d]$, $v_\Omega$ is the vector of $v$ restricted on $\Omega$. We say a vector is $k$-sparse if it has at most $k$ non-zero elements, and likewise for matrices and tensors. For a matrix $M$ of size $d_1 \times d_2$, denote by $\fronorm{M}$ its Frobenius norm and by $\nuclearnorm{M}$ its nuclear norm.
For $U\subseteq[d_1]\times[d_2]$, denote by $M_U$ the submatrix of  $M$ with entries restricted to $U$. 

We also use tensors in our algorithms to ease expressions. Note that vectors and matrices can be seen as order-$1$ and order-$2$ tensors respectively. We say that an order-$l$ tensor $A$ is symmetric if $A_{i_1, \dots, i_l} = A_{\pi(i_1, \dots, i_l)}$ for all permutations $\pi$.  Given two tensors $A$ and $B$, denote by $A\otimes B$ the outer product (or tensor product) of $A$ and $B$. We will slightly abuse $\twonorm{A}$ to denote the $L_2$-norm of a tensor $A$ by seeing it as a long vector.


\vspace{0.1in}\noindent
{\bfseries Probability.} \ 
We reserve the capital letter $G$ for a random draw from $N(\mu, \I_d)$, i.e. $G~\sim \N(\mu,\I_d)$, where $\mu\in\R^d$ is the target mean that we aim to estimate which is assumed to be $k$-sparse. Suppose that  $T$ is a finite sample set. We use $\mu_T$ to denote the sample mean of $T$, i.e. $\mu_T = \frac{1}{\abs{T}}\sum_{x \in T} x$, and use $p(T)$ to denote the random variable $p(x)$ where $x$ is drawn uniformly from $T$.



\vspace{0.1in}\noindent
{\bfseries Constants.} \ 
The capital letter $C$ and its subscript variants such as $C_1, C_2$ are used to denote positive absolute constants. However, their values may change from appearance to appearance.

\subsection{Polynomials}


Let $x = (x_1, \dots, x_d)$ be a $d$-dimensional vector in $\R^d$, and let $\ba = (\ba_1, \dots, \ba_d) \in \mathbb{N}^d$ be a $d$-dimensional multi-index. A {\em monomial} of $x$ is a product of powers of the coordinates of $x$ with natural exponents, written as $x^{\ba} := \prod_{j=1}^{d} x_j^{\ba_j}$. A {\em polynomial} of $x$, $p(x)$, is a finite sum of its monomials multiplied by real coefficients; that is, $p(x) = \sum_{\ba \in \mathcal{A}} c_{\ba} x^{\ba}$ where $\mathcal{A} \subset \mathbb{N}^d$ is a finite set of multi-indices and the $c_{\ba}$'s are real coefficients. Note that the degree of $p(x)$ is given by $\max_{\ba \in \mathcal{A}}\onenorm{\ba}$. We denote by $\mathbb{P}(\R^d, l)$ the class of polynomials on $\R^d$ with degree {at most} $l$. We will often use the probabilist's Hermite polynomials that form a complete orthogonal basis with respect to  $N(0,\I_d)$.





\begin{definition}[Hermite polynomials]\label{def:hermite}
Let $x \in \R$ be a variate. For any natural number $l \in \mathbb{N}$, the degree-$l$ Hermite polynomial is defined as $\hermite_{l}(x)=(-1)^{l}e^{\frac{x^2}{2}}\frac{d^{l}}{dx^{l}}e^{-\frac{x^2}{2}}$. For $\ba\in\mathbb{N}^d$ and $x\in\R^d$, the $d$-variate  Hermite polynomial is given by $\hermite_{\ba}(x) := \prod_{i=1}^d \hermite_{\ba_i}(x_i)$, which is of degree $\onenorm{\ba}$. 
\end{definition}


\vspace{0.1in}\noindent
{\bfseries Harmonic and homogeneous  polynomials.} \  
A polynomial $h(x) \in \mathbb{P}(\R^d, l)$ is called {\em harmonic} if it can be written as a linear combination of degree-$l$ Hermite polynomials. A polynomial $\homogeneous(x) \in \mathbb{P}(\R^d, l)$ is called {\em homogeneous} if all of its monomials have degree exactly $l$.  


\begin{fact}\label{fact:poly=tensor}
If a polynomial is degree-$l$ harmonic or homogeneous, then there is a one-to-one mapping between it and an order-$l$ symmetric tensor. 
\end{fact}

To see this, we may define an operation ``$\circ$'' such that $\hermite_l(x_i) \circ \hermite_l(x_j) = \hermite_l(x_i) \cdot \hermite_l(x_j)$ if $i \neq j$ and equals $\hermite_{2l}(x_i)$ otherwise. Then any degree-$l$ Hermite polynomial can be written as $\hermite_1(x_{i_1}) \circ \hermite_1(x_{i_2}) \dots \circ \hermite_1(x_{i_l})$ where all the indices $i_t \in [d]$. We will consider that one such sequence $(i_1, \dots, i_l)$ exactly corresponds to one degree-$l$ Hermite polynomial on $\R^d$, and there are $d^l$ number of such sequences that form all degree-$l$ Hermite polynomials. In this sense, any harmonic polynomial $h(x)$ can be written as $h(x) = \sum_{i_1, \dots, i_l} A_{i_1, \dots, i_l} \cdot \hermite_1(x_{i_1}) \circ \hermite_1(x_{i_2}) \dots \circ \hermite_1(x_{i_l})$, where $A_{i_1, \dots, i_l}$'s are the coefficients which form an order-$l$ tensor. If we choose $A$ as symmetric, it is easy to see that $A$ fully represents $h(x)$. Then, we can convert ``$\circ$'' back to the regular product by counting the number of times a particular index $j$ appearing in $(i_1, \dots, i_l)$. If we denote this number as $c_j(i_1, \dots, i_l)$, we have
\begin{equation}
h(x) = \frac{1}{\sqrt{l!}} \sum_{i_1,\dots,i_l}A_{i_1,\dots,i_l}\prod_{j} \hermite_{c_j(i_1,\dots,i_l)}(x_j) =: h_A(x), \ \text{with}\  \sum_{j=1}^{d}c_j(i_1,\dots,i_l)=l,
\end{equation}
where the factor $1/\sqrt{l!}$ is only used to normalize the magnitude of $A$ to ease our analysis.

Likewise, any homogeneous polynomial takes the form 
\begin{equation*}
\homogeneous_A(x) = \sum_{i_1,\dots,i_l}A_{i_1,\dots,i_l}\prod_{j} x_j^{c_j(i_1,\dots,i_l)}.
\end{equation*}

\vspace{0.1in}\noindent
{\bfseries Sparse polynomials.} \  
In order to define sparse polynomials, we will first specify a set of basis polynomials $\{b_1, \dots, b_{d^l}\} \subset \mathbb{P}(\R^d, l)$. In this paper, we will either choose such set as all degree-$l$ monomials or all degree-$l$ Hermite polynomials.

\begin{definition}[{$(\kappa,\psi)$-sparse polynomials}]\label{def:sparse_polynomial}
We say that a polynomial $p \in \mathbb{P}(\R^d, l)$ is $(\kappa, \psi)$-sparse if it can be represented by at most $\kappa$ number of basis polynomials and $\psi$ coordinates of the input vector. We denote by $\mathbb{P}(\R^d, l, \kappa, \psi)$ the class of ($\kappa, \psi)$-sparse polynomials.
\end{definition}

Note that when $\kappa$ and $l$ are fixed, $p(x)$ will depend on at most $\kappa \cdot l$ coordinates. Thus, the introduction of the parameter $\psi$ makes sense only when $\psi \leq \kappa \cdot l$. In our algorithm, we will always have $l \leq 2\ell$, $\kappa = 4\ell^2 k^{4\ell}$, and $\psi = 2\ell k^{2\ell}$ for some natural number $\ell \geq 1$.


%


\subsection{Representative Set and Good Set}







To ease our analysis, we will need a deterministic condition on the set of uncorrupted samples.

\begin{definition}[Representative set]\label{def:representative-set}
Given $\alpha\in(0,\frac12]$ and $\tau\in(0,1)$, we say that a sample set $\Sgood\subset \R^d$ is representative with respect to $\mathbb{P} := \mathbb{P}(\R^d, 2\ell, {4\ell^2 k^{4\ell}}, 2\ell k^{2\ell})$ if the following holds:
\begin{equation*}
\sup_{p \in \mathbb{P}}	\abs{\Pr[p(G)\geq0] - \Pr[p(\Sgood)\geq0]} \leq \epsilon_0, \ \text{where}\ \epsilon_0 := \frac{\alpha^3}{100 k^{2\ell} \cdot {\log^{2\ell}(\frac{\ell d}{\alpha \tau})} }.
\end{equation*}
\end{definition}


We show that a sufficiently large set drawn independently from $N(\mu, \I_d)$ is representative. The proof follows from the classic VC theory, and is deferred to Appendix~\ref{sec:app:sample-comp}. 


\begin{proposition}[Sample complexity]\label{prop:sample_complexity}
Given $\alpha\in(0,\frac12]$ and $\tau\in(0,1)$, let $\Sgood$ be a set consisting of $\abs{\Sgood} = C \cdot \frac{l \cdot \kappa^2 \psi \log^3 d}{\epsilon^2} \log\frac{l \cdot \kappa \psi}{\epsilon \tau}$ independent samples from $N(\mu, \I_d)$ where $C > 0$ is a sufficiently large absolute constant. Then, with probability $1-\tau$,
\begin{equation*}
\sup_{p \in \sparseP}\abs{\Pr[p(G)\geq 0] - \Pr[p(\Sgood)\geq 0]} \leq \epsilon.
\end{equation*}
In particular, when $l = 2\ell$, $\kappa = {4\ell^2 k^{4\ell}}$, $\psi=2\ell k^{2\ell}$, and $\epsilon = \frac{\alpha^3}{100 k^{2\ell} \cdot \log^{2\ell}(\frac{\ell d}{\alpha \tau}) }$ for some natural number $\ell \geq 1$, it suffices to pick $\abs{\Sgood} = C' \cdot \frac{  \ell^7 \cdot k^{14\ell}}{\alpha^6} \cdot \log^{8\ell}(\frac{ \ell d}{\alpha \tau})$ for some sufficiently large constant $C'$ so that $\Sgood$ is a representative set.
\end{proposition}



Our algorithm will progressively remove samples from $T$, and a key property that ensures the success of the algorithm is that most corrupted samples are eliminated while almost all uncorrupted samples are retained. Alternatively, we hope that $T$ contains a representative set that contributes to a nontrivial fraction. For technical reasons, we also require that all samples in $T$ lie in a small $L_\infty$-ball. 

\begin{definition}[$\alpha$-good set]\label{def:good-set}
A multiset $T\subset\R^d$ is $\alpha$-good if the following holds:
\begin{enumerate}
	\item There exists a set $\Sgood$ which is representative and satisfies {$\abs{\Sgood\cap T}\geq \max\{(1-\alpha/6)\abs{\Sgood},\alpha\abs{T}\}$}.
\item $\max_{x,y\in T} \infnorm{x-y} \leq C \cdot \sqrt{\log(d\abs{\Sgood}/\tau)}$ for some constant $C>0$. 
\end{enumerate}
\end{definition}

It is not hard to verify that the initial sample set $T$ satisfies the first condition, and will also fulfill the second one with a simple data pre-processing, as stated in the next section.


\subsection{Clustering for the Initial List}\label{sec:cluster}

Since the corrupted samples may behave adversarially, we will perform a preliminary step of clustering which splits $T$ into an initial list of subsets, among which at least one is $\alpha$-good in the sense of Definition~\ref{def:good-set}. We first show that all Gaussian samples  have bounded $L_\infty$-norm with high probability, which simply follows from the Gaussian tail bound.



\begin{lemma}\label{lem:bounded_infnorm}
Given  $\tau \in (0, 1)$, 
with probability $1-{\tau}$, we have  $\max_{x \in \Sgood} \infnorm{x-\mu}\leq\sqrt{2\log\frac{d\abs{\Sgood}}{\tau}}$, where $\Sgood$ is a set of samples drawn independently from $N(\mu, \I_d)$.
\end{lemma}





The above observation implies that for any $x, y \in \Sgood$, their distance under the $L_{\infty}$-norm metric is at most $2\sqrt{2\log(d \abs{\Sgood}/\tau)} \leq O\big(\sqrt{\ell\cdot \log\frac{\ell d}{\alpha \tau} }\big)$ as far as the size of $\Sgood$ has the same order with the one in Proposition~\ref{prop:sample_complexity}. To guarantee the existence of such $\Sgood$, it suffices to draw a corrupted sample set $T$ that is ${1}/{\alpha}$ times larger than $\abs{\Sgood}$. The lemma below further shows that this is sufficient to guarantee the existence of an $\alpha$-good subset of $T$.

\begin{algorithm}[t]
\caption{$\cluster(T, \alpha, \tau, \ell)$}
\label{alg:cluster}
\begin{algorithmic}[1]
\REQUIRE A multiset of samples $T\subset \R^d$, parameter $\alpha\in(0,1/2]$, failure probability $\tau \in (0, 1)$, degree of polynomials $\ell \geq 1$.

\STATE  A set of centers $\mathcal{C}\assign\emptyset$, radius $\gamma\assign C_0 \cdot \sqrt{\ell \cdot \log\frac{\ell d}{\alpha\tau} }$ for some constant $C_0 > 0$.

\STATE  For each $x\in T$, proceed as follows: {\bfseries if} there are at least $\alpha\cdot\abs{T}$ samples $y$ in $T$ that satisfy  $\infnorm{x-y} \leq 2\gamma$,
and no sample $x'\in\mathcal{C}$ satisfies $\infnorm{x-x'} \leq 6\gamma$ {\bfseries then} $\mathcal{C}\assign\mathcal{C}\cup \{ x \}$.

\STATE For each $x_i\in\mathcal{C}$, let $T_i = \{y\in T: \infnorm{x_i-y} \leq 6\gamma\}$.

\RETURN $\{T_1, \dots, T_{\abs{\mathcal{C}}} \}$.

\end{algorithmic}
\end{algorithm}

\begin{lemma}[\cluster]\label{lem:cluster}
Given $\alpha\in(0,\frac12]$ and $\tau\in(0,1)$, let $T$ be the sample set given to the learner. If $\abs{T} = C \cdot \frac{  \ell^7 \cdot k^{14\ell}}{\alpha^7} \cdot \log^{8\ell}(\frac{ \ell d}{\alpha \tau})$ and a $(2\alpha)$-fraction are independent samples from $N(\mu, \I_d)$, Algorithm~\ref{alg:cluster} returns a list of at most ${1}/{\alpha}$ many subsets of $T$, such that with probability at least $1-\tau$, at least one of them is an $\alpha$-good set.
\end{lemma}


{As will be clear in our analysis,} the motivation of bounding the $L_\infty$-distance is to make sure that the function value of any $p(x)=h_A(x-\mu_T) \in \mathbb{P}(\R^d, l, \kappa, \psi)$ is bounded for samples in the $\alpha$-good subset $T_i$. This is because when there exist a significant fraction of good samples in $T_i$, we want to efficiently distinguish the corrupted and uncorrupted ones. A value-bounded polynomial function will facilitate our analysis on the function variance. 

\begin{lemma}\label{lem:function_value_bound}
Suppose that $T$ is $\alpha$-good and a polynomial $p \in \mathbb{P}(\R^d, l, 4\ell^2 k^{4\ell}, 2\ell k^{2\ell})$ satisfies the following: there exists a symmetric order-$l$ tensor $A$ such that $\twonorm{A}\leq1$ and $p(x)=h_A(x- \mu_T)$. Then, it holds that $\max_{x,y\in T}\abs{p(x)-p(y)} \leq 2k^{\ell}\cdot \gamma^l$, where $\gamma=C_0 \cdot\sqrt{\ell \cdot \log(\frac{ \ell d}{\alpha \tau})}$.
\end{lemma}

%
%

%% file: algorithm.tex
\section{Main Algorithms and Performance Guarantees}\label{sec:algorithm}

We start with a review of the multifiltering framework that has been broadly used in prior works~\cite{dia2018list,dia2020list,dia2021cluster}, followed by a highlight of our new techniques.



The multifiltering framework, i.e. Algorithm~\ref{alg:listdecodable}, includes three major steps. The first step is to invoke \cluster (Algorithm~\ref{alg:cluster}) to generate an initial list $\calL$ which guarantees the existence of an $\alpha$-good subset of $T$ (see Lemma~\ref{lem:cluster}). We then imagine that there is a tree with root being the original contaminated sample set $T$ and each child node of the root represents a member in $\calL$.
The algorithm  iterates through these child nodes and performs one of the following: (1) creating a leaf node which is an estimate of the target mean; (2)  creating one or two child nodes where are subsets of the parent node; (3) certifying that the set cannot be $\alpha$-good and delete branch. In the end, if all leaves of the tree cannot be further split or deleted, the mean of the subsets on leaf nodes will be collected as a list $M$. {It is worth noting that the goal of algorithmic design is to guarantee that there always exists a branch that includes only $\alpha$-good subsets. In other words, at any level of the algorithm, at least one of the subsets of $T$ is $\alpha$-good, which ensures the existence of a good estimation in the returned list $M$.} The final step is a black-box  algorithm that reduces the  size of $M$ from $O(\poly(1/\alpha))$ to $O(1/\alpha)$, which is due to \cite{dia2018list}. Our technical contributions lie into an attribute-efficient implementation of the first and second steps. In this section, we elaborate on the second step, i.e. the \aefilter algorithm.

\begin{algorithm}[t]
\caption{Main Algorithm: Attribute-Efficient List-Decodable Mean Estimation}
\label{alg:listdecodable}
\begin{algorithmic}[1]
\REQUIRE A multiset of samples $T\subset \R^d$, parameter $\alpha\in(0,1/2]$, failure probability $\tau \in (0, 1)$, degree of polynomials $\ell \geq 1$.

\STATE $\{T_1, \dots, T_m\} \gets \cluster(T, \alpha, \tau, \ell)$, $\calL \gets \{ (T_1, \alpha/2), \dots, (T_m, \alpha/2)\}$, $M \gets \emptyset$.

\WHILE{ $\calL \neq \emptyset$}

\STATE $(T',\alpha') \gets$ an element in $\calL$, $\calL \gets \calL \backslash \{(T', \alpha')\}$.

\STATE $\ANS \gets \aefilter(T', \alpha', \tau/\abs{T}, \ell)$.\label{step:main-aemf}

\begin{enumerate}[label=(\roman*)]
\item  {\bfseries  if} $\ANS$ is a vector {\bfseries  then} add it into $M$.

\item {\bfseries  if} $\ANS$ is a list of $(T_i,\alpha_i)$ {\bfseries  then} append those with $\alpha_i \leq 1$ to $\calL$.

\item {\bfseries  if} $\ANS = \NO$ {\bfseries  then} go to the next iteration.
\end{enumerate}

\ENDWHILE

\RETURN $\listred(T,\alpha,\ell, M)$. 
\end{algorithmic}
\end{algorithm}


\begin{algorithm}[t]
\caption{$\aefilter(T, \alpha, \tau, \ell)$}
\label{alg:mainSub-highdegree}
\begin{algorithmic}[1]
\REQUIRE A multiset of samples $T\subset \R^d$, parameter $\alpha\in(0,1/2]$, failure probability $\tau \in (0, 1)$, degree of polynomials $\ell \geq 1$.

\STATE $\tilde{\Sigma} \gets \E[P_{d,\ell}(T-{\mu_T}) \cdot P_{d,\ell}(T-{\mu_T})^{\top}]$,  and $P_{d,\ell}(x)$ is the column vector of all degree-$\ell$ Hermite polynomials of $x$. 


\STATE  $\{(i_t, j_t)\}_{t\geq 1}^{\frac{1}{2}(k^{2\ell} + k^{\ell})} \gets$ index set of the $k^{\ell}$  diagonal entries and $\frac12(k^{2\ell}-k^{\ell})$ entries above the main diagonal of $\tilde{\Sigma}$ with largest magnitude. $U \gets \{(i_t, j_t)\}_{t\geq 1} \cup \{(j_t, i_t)\}_{t\geq 1}$, $U' \gets I \times I$, with $I = \{i_t\}_{t \geq 1} \cup \{j_t\}_{t\geq 1}$.

\STATE $\lambdasparse \gets {\big[C_1\cdot(\ell+ C_1 \log\frac{1}{\alpha}) \cdot\log^2(2+\log\frac{1}{\alpha}) \big]^{2\ell} }$ for large enough constant $C_1 > 0$.



\STATE {\bfseries if} $\fronorm{(\tilde{\Sigma})_{U}} \leq \lambdasparse$ {\bfseries  then return} $\hat{\mu} \gets \hard_k(\mu_T)$.\label{step:MF-return-mean}


\STATE $(\lambda^*, v^*) \gets$ the largest eigenvalue and eigenvector of $(\tilde{\Sigma})_{U'}$. \label{step:MF-pca}

\IF{$\lambda^* \geq \lambdasparse$} \label{step:MF-linear-condition}


\STATE {\bfseries if} $\ell = 1$ {\bfseries then} $\ANS \gets \bmf(T, \alpha, \tau, p_1)$ {\bfseries else} $\ANS \gets \harmf(T, \alpha, \tau, p_1)$ where $p_1(x) := v^*\cdot P_{d,\ell}(x - \mu_T)$. \label{step:MF-linear} 


\ELSE 

\STATE $p_2(x) \gets \frac{1}{\fronorm{A'}}\cdot{\(P_{d,\ell}(x - \mu_T)^{\top} \cdot A' \cdot P_{d,\ell}(x - \mu_T)\)}$ with $A' :=(\tilde{\Sigma})_{U'}$.

\STATE $\ANS \gets \harmf(T, \alpha, \tau, p_2)$. \label{step:MF-quad} 


\ENDIF

\RETURN $\ANS$.
\end{algorithmic}
\end{algorithm}

\subsection{Overview of Attribute-Efficient Multifiltering}\label{subsubsec:tech}

The \aefilter algorithm is presented in Algorithm~\ref{alg:mainSub-highdegree}. The starting point of the algorithm is a well-known fact that if the adversary were to significantly deteriorate our estimate on $\mu$, the spectral norm of a certain sample covariance matrix $\tilde{\Sigma}$ would become large  \cite{diakonikolas2016robust,lai2016agnostic}. In order to achieve attribute-efficient sample complexity $O(\poly(k, \log d))$, it is however vital to control the spectral norm only on $k^{\ell}$-sparse directions for some pre-specified polynomial degree $\ell \geq 1$, which can further be certified by a small Frobenius norm restricted on the largest $k^{2\ell}$ entries. If the restricted Frobenius norm is sufficiently small, it implies that the sample covariance matrix behaves as a Gaussian one, and the algorithm returns the empirical mean truncated to be $k$-sparse (see Step~\ref{step:MF-return-mean}). Otherwise, the algorithm will invoke either \bmf (i.e. Algorithm~\ref{alg:basic}) or \harmf (i.e. Algortihm~\ref{alg:hmf}) to examine the concentration of a polynomial of the empirical data to that of Gaussian. Both algorithms will either assert that the current sample set does not contain a sufficiently large  amount of Gaussian samples, or will prune many corrupted samples to increase the fraction of Gaussian ones. A more detailed description of the two algorithms can be found in Section~\ref{subsec:basic} and Section~\ref{subsec:hmf} respectively. What is subtle in Algorithm~\ref{alg:mainSub-highdegree} is that we will check the maximum eigenvalue $\lambda^*$ of the empirical covariance matrix $\tilde{\Sigma}$ restricted on a carefully chosen subset $U'$, which corresponds to the maximum eigenvalue on a certain $(2k^{2\ell})$-sparse direction. If $\lambda^*$ is too large, this indicates an {\em easy} problem since it must be the case that the adversary corrupted the samples in an aggressively way. Therefore, it suffices to prune outliers using a degree-$\ell$ polynomial $p_1$ which is simply the projection of $P_{d,\ell}(x-\mu_T)$ onto the span of the maximum eigenvector; see Step~\ref{step:MF-linear} in Algorithm~\ref{alg:mainSub-highdegree}. On the other hand, if $\lambda^*$ is on a moderate scale, it indicates that the adversary corrupted the samples in a very delicate way so that it passes the tests of both Frobenius norm and spectral norm. Now the main idea is to check the concentration of higher degree polynomials induced by the sample set; we show that it suffices to construct a degree-$2\ell$ harmonic polynomial; see Step~\ref{step:MF-quad}.

While sparse mean estimation has been studied in \cite{dia2019outlier} and the idea of using restricted Frobenius norm and filtering was also developed, we note that their analysis only holds in the mild corruption regime where $\alpha > 1/2$. To establish the main results, we will leverage the tools from \cite{dia2018list}, with a specific treatment on the fact that $\mu$ is $k$-sparse, to ensure an attribute-efficient sample complexity bound. As we will show later, a key idea to this end is to utilize a sequence of carefully chosen sparse polynomials in the sense of Definition~\ref{def:sparse_polynomial} along with sparsity-induced filters.


The performance guarantee of \aefilter is as follows.
\begin{theorem}[Algorithm~\ref{alg:mainSub-highdegree}]\label{thm:mainsub}
Consider Algorithm~\ref{alg:mainSub-highdegree} and denote by $\ANS$ its output. With probability $1-\tau$, the following holds. $\ANS$ cannot be $\TBD$. If $\ANS$ is a $k$-sparse vector and if $T$ is $\alpha$-good, then $\twonorm{\mu-\hat{\mu}}\leq {\tilde{O}\big(\alpha^{-\frac{1}{2\ell}}\sqrt{\ell}(\ell+\log\frac{1}{\alpha}) \big)}$. If $\ANS = \NO$, then $T$ is not $\alpha$-good. If $\ANS = \{(T_i, \alpha_i)\}_{i=1}^m$ for some $m \leq 2$, then $T_i \subset T$ for all $i \in [m]$ and $\sum_{i=1}^{m} \frac{1}{\alpha_i^2} \leq \frac{1}{\alpha^2}$; if additionally $T$ is $\alpha$-good, then at least one $T_i$ is $\alpha_i$-good. Finally, the algorithm runs in time $O\big(\poly(\abs{T}, d^{\ell})\big)$.
\end{theorem}

\subsection{Analysis of \aefilter}\label{subsec:aemf}

We first show that if the restricted Frobenius norm of the sample covariance matrix is small, then the sample mean is a good estimate of the target mean.

\begin{lemma}\label{lem:returned_vector}
Consider Algorithm~\ref{alg:mainSub-highdegree}. If the algorithm returns a vector $\hat{\mu}$ at Step~\ref{step:MF-return-mean} and if $T$ is $\alpha$-good, we have that $\twonorm{\hat{\mu}-\mu}\leq {O\big(\alpha^{-\frac{1}{2\ell}}\sqrt{\ell} \cdot (\ell+\log\frac{1}{\alpha}) \cdot \log^2(2+\log\frac{1}{\alpha})\big) }$.
\end{lemma}

Next, we give performance guarantees on the remaining steps of Algorithm~\ref{alg:mainSub-highdegree}, where we consider the case that the algorithm does not return at Step~\ref{step:MF-return-mean}. Namely, the algorithm will either reach at Step~\ref{step:MF-linear} or Step~\ref{step:MF-quad}, and will return the $\ANS$ obtained thereof. These two steps will invoke \bmf or \harmf on different sparse polynomials. Observe that both algorithms may return 1) ``$\NO$'', which certifies that the current input set $T$ is not $\alpha$-good;  2) a list of subsets $\{(T_i, \alpha_i)\}_{i=1}^m$ for some $m \leq 2$, on which Algorithm~\ref{alg:mainSub-highdegree} will be called in a recursive manner; or 3) $\TBD$, which indicates that the algorithm is uncertain on $T$ being $\alpha$-good. In the following, we prove that the way that we invoke \bmf and \harmf ensures that they will never return $\TBD$ when being called within Algorithm~\ref{alg:mainSub-highdegree}. We then give performance guarantees on these two filtering algorithms when they return ``$\NO$'' or $\{(T_i, \alpha_i)\}_{i=1}^m$, thus establishing Theorem~\ref{thm:mainsub}.

Let us consider that the algorithm reaches Step~\ref{step:MF-linear}, i.e. the largest eigenvalue on one sparse direction is larger than the threshold $\lambdasparse$. It is easy to see that when $\ell = 1$, $\ANS$ cannot be $\TBD$ since the only way that \bmf returns $\TBD$ is when $\Var[p(T)]$ is not too large, but this would violate the condition that $\lambda^* > \lambdasparse$ in view of our setting on $\lambdasparse$. Similarly, we show that under the large $\lambda^*$ regime, \harmf will not return $\TBD$ either. Thus, we have the following lemma.

\begin{lemma}\label{lem:linear_filter}
Consider Algorithm~\ref{alg:mainSub-highdegree}. If it reaches Step~\ref{step:MF-linear}, then $\ANS \neq \TBD$.
\end{lemma}

Now it remains to consider the case that the algorithm reaches Step~\ref{step:MF-quad}, which is more subtle since the evidence from the magnitude of the largest restricted eigenvalue is not so strong to prune outliers. Note that this could happen even when $T$ contains many outliers, since $\lambda^*$ is not the maximum eigenvalue on all sparse directions but on a submatrix indexed by $U'$. Fortunately, if $\lambda^*$ is not large, we show that the algorithm can still make progress by calling \harmf on {degree-$2\ell$} sparse polynomials. This is because higher-degree polynomials are more sensitive to  outliers than low-degree polynomials, as far as we can certify the concentration of high-degree polynomials on clean samples. As a result, we will have the following guarantee.

\begin{lemma}\label{lem:quadratic_filter}
Consider Algorithm~\ref{alg:mainSub-highdegree}. If it reaches  Step~\ref{step:MF-quad}, then $\ANS \neq \TBD$.
\end{lemma}




\subsubsection{Basic Multifilter for Sparse Polynomials} \label{subsec:basic}

\begin{algorithm}[t]
\caption{$\bmf(T, \alpha, \tau, p)$}
\label{alg:basic}
\begin{algorithmic}[1]
\REQUIRE A multiset of samples $T\subset \R^d$, parameter $\alpha\in(0,1/2]$, failure probability $\tau \in (0, 1)$, a  polynomial $p \in \mathbb{P}(\R^d, l, 4\ell^2 k^{4\ell}, 2\ell k^{2\ell})$ such that $l \leq 2\ell$, $\Var[p(G)]\leq1$, and $p(x) = h_A(x - \mu_T)$.


\STATE  $R\assign {(C_1\cdot\log\frac{1}{\alpha})}^{l/2}$, $\gamma\assign C_0 \cdot \sqrt{\ell \cdot \log\frac{\ell d}{\alpha\tau} }$.

\STATE  {{\bfseries if} $\max_{x,y\in T}\abs{p(x)-p(y)} > {2 k^{\ell}\cdot \gamma^{l}}$ {\bfseries  then return} ``$\NO$''.}\label{step:BMF-return-no}

\IF{there is an interval $[a, b]$ of length $C_1\cdot R\cdot\log(2+\log\frac{1}{\alpha})$ that contains at least $(1- \frac{\alpha}{2})$-fraction of samples in $\{p(x): x\in T \}$} \label{step:BMF-find-interval}

\IF{$\Var[p(T)] \leq C_1\cdot {\big(\ell+C_1\log\frac{1}{\alpha}\big)}^l \cdot \log^2(2+\log\frac{1}{\alpha})$}

\RETURN ``$\TBD$''. \label{step:BMF-return-yes}

\ELSE

\STATE Find a threshold $t>2R$ such that
\begin{equation*}
\Pr_{x\sim T}\big[\min\{\abs{p(x)-a},\abs{p(x)-b}\} \geq t \big] > \frac{32}{\alpha}\exp(-(t-2R)^{2/l}) + {\frac{2\alpha^2}{ k^{2\ell} \log^{l}(\frac{\ell d}{\alpha \tau})} }.
\label{eq:oneCluster}
\end{equation*} \label{step:BMF-onecluster-threshold}
\STATE $T' \assign \{x\in T: \min\{\abs{p(x)-a},\abs{p(x)-b}\}\leq t\}$, $\alpha' \assign \alpha\cdot\(\frac{(1-\alpha/8)\abs{T}}{\abs{T'}} + \frac{\alpha}{8}\)$. \label{step:BMF-onecluster-filter}
\RETURN $\{(T',\alpha')\}$.\label{step:BMF-return-onecluster}
\ENDIF

\ELSE

\STATE  Find $t\in\R$, $R' >0$ such that the sets $T_1 :=\{x\in T: p(x)>t-R'\}$ and $T_2 :=\{x\in T: p(x)<t+R'\}$ satisfy
\begin{equation*}
\abs{T_1}^2 + \abs{T_2}^2 \leq \abs{T}^2 (1-\alpha/100)^2\ \text{and }\abs{T} - \max(\abs{T_1},\abs{T_2}) \geq \alpha\abs{T}/4.
\end{equation*} \label{step:BMF-twocluster-threshold}

\STATE  $\alpha_i \gets \alpha\cdot(1-\alpha^2/100)\cdot \abs{T}/\abs{T_i}$, for $i=1,2$.

\RETURN $\{(T_1,\alpha_1),(T_2,\alpha_2)\}$.\label{step:BMF-return-twocluster}

\ENDIF

\end{algorithmic}
\end{algorithm}

The \bmf algorithm (Algorithm~\ref{alg:basic}) is a key ingredient in the multifiltering framework. It takes as input a sparse polynomial $p$ and uses it to certify whether $T$ is $\alpha$-good and sufficiently concentrated. The central idea is to measure how $p(T)$ distributed and compare it to that of the distribution of $p(G)$. We require the input $p$ has certifiable variance on $G$, i.e. $\Var[p(G)]\leq1$, as otherwise, it could filter away a large number of the good samples. We note that the bounded variance condition is always satisfied for degree-$1$ Hermite polynomials under proper normalization, while for high-degree polynomials, one cannot invoke \bmf directly (see Section~\ref{subsec:hmf} for a remedy).

The way that \bmf certifies the input sample set $T$ not being $\alpha$-good is quite simple: if not all samples lie in a small $L_{\infty}$-ball, it returns ``$\NO$'' at Step~\ref{step:BMF-return-no}, in that this contradicts Lemma~\ref{lem:function_value_bound}. Otherwise, the algorithm will attempt to search for a finer interval $[a, b]$ such that it includes most of the samples. If such interval exists, then either the adversary corrupted the samples such that the sample variance is as small as that of Gaussian while the sample mean may deviate far from the target, in which case \bmf returns $\TBD$ at Step~\ref{step:BMF-return-yes}; or the sample variance is large, in which case it is possible to construct a sparsity-induced filter to prune outliers (see Steps~\ref{step:BMF-onecluster-threshold} and \ref{step:BMF-onecluster-filter}). We note that in Step~\ref{step:BMF-onecluster-threshold}, the first term on the right-hand side is derived from Chernoff bound for degree-$l$ Gaussian polynomials and the second term is due to concentration of empirical samples to Gaussian (see Definition~\ref{def:representative-set}), both of which are scaled by a factor $8/\alpha$ so that the number of the samples removed from $T$ is ${8}/{\alpha}$ times more than that of the good samples in the representative set $\Sgood \subset T$, which means most of the removed samples are outliers. We show by contradiction the existence of the threshold $t$ (see Lemma~\ref{lem:threshold_for_one}). In fact, had such threshold $t$ not existed, the set $T$ must be sufficiently concentrated such that the algorithm would have returned at Step~\ref{step:BMF-return-yes}. This essentially relies on our result of the initial clustering of Algorithm~\ref{alg:cluster}, which guarantees that each subset $T$ is bounded in a small $L_\infty$-ball and the function value of $p$ on the $\alpha$-good $T$ does not change drastically (Lemma~\ref{lem:function_value_bound}). We then show that equipped with such threshold $t$, $T'$ is a subset of $T$ and it is $\alpha'$-good if $T$ is $\alpha$-good (Lemma~\ref{lem:T'-good}).


When there is no such short interval $[a,b]$, the algorithm splits $T$ into two overlapping subsets $\{T_1,T_2\}$ such that $T_1\cap T_2$ is large enough to contain most of the samples in $\Sgood$. This guarantees that most of the samples in $\Sgood$ (if $T$ is $\alpha$-good) are always contained in one subset and thus there always exists an $\alpha$-good subset of $T$. We show that an appropriate threshold $t$ can also be found at Step~\ref{step:BMF-twocluster-threshold} (Lemma~\ref{lem:threshold_for_two}), and at least one $T_i$ is $\alpha_i$-good if $T$ is $\alpha$-good. 

As a result, we have the following guarantees for Algorithm~\ref{alg:basic}; see Appendix~\ref{sec:app:proof-main} for the full proof.

\begin{theorem}[\bmf]\label{thm:bmf}
Consider Algorithm~\ref{alg:basic}. Denote by $\ANS$ its return. Suppose that $T$ being $\alpha$-good implies $\Var[p(G)] \leq 1$. Then with probability $1-\tau$, the following holds. $\ANS$ is either ``$\NO$'', ``$\TBD$'', or a list of $\{(T_i, \alpha_i)\}_{i=1}^m$ with $m \leq 2$. 1) If $\ANS =\NO$, then $T$ is not $\alpha$-good. 2) If $\ANS = \TBD$, then $\Var[p(T)] \leq O\big({(\ell+\log\frac{1}{\alpha})}^l \cdot\log^2(2+\log\frac{1}{\alpha})\big)$; and if additionally $T$ is $\alpha$-good, then $\abs{\E[p(G)] - \E[p(T)]} \leq O\big({(\ell+\log\frac{1}{\alpha})}^{\frac{l}{2}}\cdot\log(2+\log\frac{1}{\alpha})\big)$. 3) If $\ANS=\{(T_i, \alpha_i)\}_{i=1}^m$, then $T_i \subset T$ and $\sum_{i} \frac{1}{\alpha_i^2} \leq \frac{1}{\alpha^2}$ for all $i \in [m]$; if additionally $T$ is $\alpha$-good, then at least one $T_i$ is $\alpha_i$-good.
\end{theorem}


\begin{algorithm}[t]
\caption{$\harmf(T, \alpha, \tau, p)$}
\label{alg:hmf}
\begin{algorithmic}[1]
\REQUIRE A multiset of samples $T\subset \R^d$, parameter $\alpha\in(0,1/2]$, failure probability $\tau \in (0, 1)$, a  polynomial $p \in \mathbb{P}(\R^d, l, 2\ell k^{2\ell}, 2\ell k^{2\ell})$ such that $p(x) = h_A(x - \mu_T)$ and $\twonorm{A}=1$.



\FOR{ $l'=0, 1, \dots, l$}

\STATE Let $B^{(l')}$ be an order-$2l'$ tensor with 
\begin{equation*}
B^{(l')}_{i_1,\dots,i_{l'},j_1,\dots,j_{l'}} = \sum_{k_{l'+1},\dots,k_l} A_{i_1\dots,i_{l'},k_{l'+1},\dots,k_l}A_{j_1\dots,j_{l'},k_{l'+1},\dots,k_l }.
\end{equation*}

\STATE Consider $B^{(l')}$ as a $d^{l'}\otimes d^{l'}$ symmetric matrix by grouping each of the $i_1,\dots,i_{l'}$ and $j_1,\dots,j_{l'}$ coordinates together. Apply eigenvalue decomposition on $B^{(l')}$ to obtain $B^{(l')} = \sum_{i} \lambda_i V_i \otimes V_i$.\label{step:HMF-calculate}

%

\STATE $\ANS_i \gets \mmf(T,V_i,l',\alpha,\tau/(ld^l))$ for every $V_i$. If $\ANS_i = \NO$ or a list of $\{(T_j,\alpha_j)\}$ for some $i$, then {\bfseries return} $\ANS_i$. If $\ANS_i = \TBD$, {\bf continue}. \label{step:HMF-callMMF}

\ENDFOR

\STATE $\ANS \gets \bmf(T, \alpha, \tau, \frac{1}{\beta}h_A(x- \mu_T))$ with $\beta := {\(C_1\cdot(1+\log\frac{1}{\alpha})\cdot\log^2(2+\log\frac{1}{\alpha})\)^{\frac{l}{2}} } $.  If $\ANS = \NO$ or a list of $(T_j,\alpha_j)$, {\bfseries return} $\ANS$. If $\ANS = \TBD$, still {\bfseries return} ``$\NO$''. \label{step:HMF-certified}


\end{algorithmic}
\end{algorithm}

\subsubsection{Harmonic Multifilter with Hermite Polynomials}\label{subsec:hmf}


Recall that applying \bmf (Algorithm~\ref{alg:basic}) on a polynomial $p$ requires $\Var[p(G)]\leq1$. It is nontrivial to verify this condition for a high-degree polynomial $p$, as the variance of high-degree Gaussian polynomials  depends on the distribution mean, i.e. $\mu-{\mu_T}$ in this case, which is unfortunately unknown. As a remedy, notice that for any harmonic polynomial $h_A(x)$, $\E_{x\sim\N(\mu',\I)}[h_A(x)^2]$ equals the summation of homogeneous polynomials of $\mu'$, which can also be seen as the expectation of multilinear polynomials over independent variables $X_{(i)}\sim\N(\mu',\I_d)$. Thus, we only need to verify the expectation of these corresponding multilinear polynomials, whose variance on $G$ does not hinge on $\mu'$. The harmonic multifilter is presented in Algorithm~\ref{alg:hmf}, where the subroutine \mmf can be found in Appendix~\ref{sec:app:omit-alg}.
We first present the guarantee when Algorithm~\ref{alg:hmf} returns all $\TBD$ at Step~\ref{step:HMF-callMMF} and reaches Step~\ref{step:HMF-certified}, where we can certify a bounded variance for $h_A(x-\mu_T)$ on $G$.

\begin{lemma}[Variance of $p$]\label{lem:variance_G}
Consider Algorithm~\ref{alg:hmf}. If it reaches Step~\ref{step:HMF-certified} and $T$ is $\alpha$-good, then we have $\E[h_A(G-\mu_T)^2] \leq \beta^2$.
\end{lemma}

Based on Lemma~\ref{lem:variance_G}, we have that $\Var[h_A(G-\mu_T)/\beta]\leq1$, for which we can invoke \bmf on  $h_A(x-\mu_T)/\beta$ and Theorem~\ref{thm:bmf} can be applied immediately. 
We are ready to elaborate the proof ideas for Lemma~\ref{lem:linear_filter} and \ref{lem:quadratic_filter}.
First, observe that \bmf returns ``$\TBD$'' at Step~\ref{step:HMF-certified} if and only if $\Var[h_A(T-\mu_T)/\beta] \leq C_1\cdot {\big(\ell+C_1\log\frac{1}{\alpha}\big)}^l \cdot \log^2(2+\log\frac{1}{\alpha})$. 
Now return to Algorithm~\ref{alg:mainSub-highdegree}. When $h_A(x-\mu_T)=p_1(x)$, this could not happen because 
$\Var[p_1(T)] 
= \Var[v^*\cdot P_{d,\ell}(T - \mu_T)] 
\geq \lambda^* \geq \lambdasparse ={\big[C_1\cdot(\ell+ C_1 \log\frac{1}{\alpha}) \cdot\log^2(2+\log\frac{1}{\alpha}) \big]^{2\ell} } \geq \beta^2 \cdot C_1\cdot {\big(\ell+C_1\log\frac{1}{\alpha}\big)}^{\ell} \cdot \log^2(2+\log\frac{1}{\alpha})$. A contradiction that implies Lemma~\ref{lem:linear_filter}.
When $h_A(x-\mu_T)=p_2(x)$, the case is more delicate. Here, we instead show that $T$ must not be $\alpha$-good and \harmf will return ``$\NO$'' correctly. This is because if $T$ is $\alpha$-good, Proposition~\ref{thm:bmf} implies that $\E[p_2(G)]$ is close to $\E[p_2(T)]$, and together with Lemma~\ref{lem:variance_G} we can show that $\E[p_2(T)]$ is small. However, by construction $\big\lVert(\tilde{\Sigma})_U\big\rVert = \E[p_2(T)]\geq\lambdasparse$, a contradiction that gives Lemma~\ref{lem:quadratic_filter}. The detailed proof can be found in Appendix~\ref{subsec:app:proof-hmf}.

%% file: conclusion.tex
\section{Conclusion and Future Work}\label{sec:conclusion}

In this paper, we developed an attribute-efficient mean estimation algorithm which achieves sample complexity poly-logarithmic in the dimension with low-degree sparse polynomials under the list-decodable setting.  A natural question is whether the current techniques could be utilized to attribute-efficiently solve the other list-decodable  problems, such as learning of halfspaces and linear regression. 

%% file: appendix.tex
\section{Omitted Proofs from Section~\ref{sec:pre}}
\label{sec:app:proof-pre}

\subsection{Proof of Proposition~\ref{prop:sample_complexity}}\label{sec:app:sample-comp}



\begin{proof}
Fix a subset $\Omega \subset [d]$ with size $\psi$, and then fix a set of $\kappa$ monomials on $\Omega$ with degree at most $l$, denoted by $\mathcal{M}(\Omega, l)$. Let $\mathbb{P}(\R^d, \mathcal{M}(\Omega, l), \Omega)$ be the induced class of polynomials. Note that $\sparseP = \cup_{\Omega} \cup_{\mathcal{M}(\Omega, l)}  \mathbb{P}(\R^d, \mathcal{M}(\Omega, l), \Omega)$. 

It is easy to see that for any $p \in \mathbb{P}(\R^d, \mathcal{M}(\Omega, l), \Omega)$, it can be represented by a linear combinations of the $\kappa$ monomials. Thus, the VC dimension of this class equals $\kappa + 1$. Then, we note that there are $\sum_{j=0}^{\psi}\binom{d}{j}$ choices of $\Omega$, and for any given $\Omega$, there are $\sum_{j=0}^{\kappa}\binom{2d^{l}}{j}$ choices of $\mathcal{M}(\Omega, l)$. Therefore, the total number of the subclass $\mathbb{P}(\R^d, \mathcal{M}(\Omega, l), \Omega)$ is at most
\begin{equation}
\sum_{j=0}^{\psi}\binom{d}{j} \cdot \sum_{j=0}^{\kappa}\binom{2d^{l}}{j} \leq \( \frac{ed}{\psi} \)^{\psi} \cdot \( \frac{2e d^{l}}{\kappa} \)^{\kappa}.
\end{equation}
The concept class union argument states that for $\mathcal{H} = \cup_{i=1}^m \mathcal{H}_i$, the VC dimension of $\mathcal{H}$ is upper bounded by $O(\max\{V, \log m + V \log\frac{\log m}{V} \})$ or loosely $O(V \log m)$, where $V$ is an upper bound on the VC dimension of all $\mathcal{H}_i$. In our case, we have $V = \kappa + 1$ and $m \leq \( \frac{ed}{\psi} \)^{\psi} \cdot \( \frac{2e d^{l}}{\kappa} \)^{\kappa}$. By calculation, we can show that the VC dimension of $\mathbb{P}(\R^{d}, l, \kappa, \psi)$ is upper bounded by
\begin{equation}\label{eq:tmp:d'}
(\kappa+1) \cdot \psi  \log\frac{ed}{\psi} \cdot \kappa \log\frac{2e d^{l}}{\kappa} \leq l \cdot \kappa^2 \psi  \log^2 d =: d'.
\end{equation}
Recall that the VC theory states that for any $\epsilon, \tau \in (0, 1)$, as long as $\abs{\Sgood} \geq C \( \frac{d'}{\epsilon^2}\log \frac{d'}{\epsilon} + \frac{1}{\epsilon^2} \log\frac{1}{\tau}\)$ for some absolute constant $C > 0$, the following holds with probability $1-\tau$:
\begin{equation}
\sup_{p \in \mathbb{P}(\R^d, l, \kappa, \psi)}\abs{\Pr[p(G)\geq 0] - \Pr_{x\sim \Sgood}[p(x)\geq 0]} \leq \epsilon.
\end{equation}
With the expression of $d'$ in \eqref{eq:tmp:d'}, it is not hard to see that we can set $\abs{\Sgood} = C \cdot \frac{l \cdot \kappa^2 \psi \log^3 d}{\epsilon^2} \log\frac{l \cdot \kappa \psi}{\epsilon \tau}$ for some absolute constant $C > 0$ to ensure that the above holds.


{When $l = 2\ell$, $\kappa = {4\ell^2 k^{4\ell}}$, $\psi=2\ell k^{2\ell}$, and $\epsilon = \frac{\alpha^3}{100 k^{2\ell} \cdot \log^{2\ell}(\frac{\ell d}{\alpha \tau}) }$ for some natural number $\ell \geq 1$, by algebraic calculation, it suffices to pick $\abs{\Sgood} = C' \cdot \frac{  \ell^7 \cdot k^{14\ell}}{\alpha^6} \cdot \log^{5\ell}(\frac{ \ell d}{\alpha \tau})$ for some sufficiently large constant $C'$. This completes the proof.}
\end{proof}

\subsection{Proof of Lemma~\ref{lem:bounded_infnorm}}

\begin{proof}
By the standard tail bound of Gaussian distribution, for any $x$ drawn from $N(\mu,\I_d)$, it holds that for any given index $i \in [d]$, $\Pr[\abs{x_i-\mu_i}\geq t]\leq 2\exp(-t^2/2)$. By taking union bound over both index $i$ and sample $x \in \Sgood$, we have $\Pr[\max_{x \in \Sgood} \max_{i \in [d]}\abs{x_i-\mu_i}\geq t]\leq 2 d \abs{\Sgood}\exp(-t^2/2)$. Choosing $t=\sqrt{2\log(d\abs{\Sgood}/\tau)}$ completes the proof.
\end{proof}

\subsection{Proof of Lemma~\ref{lem:cluster}}

\begin{proof}
Let $\Sgood$ be the subset of $T$ containing the samples drawn i.i.d. from $N(\mu, \I_d)$. Since $\abs{\Sgood}=2\alpha\cdot\abs{T}$, we know that $\Sgood$ is a representative set with probability at least $1-\tau$ in light of Prop.~\ref{prop:sample_complexity}.

Consider Algorithm~\ref{alg:cluster}. If for all $x, y\in T$, we have
\begin{equation}\label{eq:cluster_diameter}
\infnorm{x-y} \leq 6 \gamma,
\end{equation}
then the algorithm returns only one cluster and the lemma follows immediately.

If that is not the case, we first note that, with probability at least $1-\tau$ all of the samples in $\Sgood$ satisfy Eq.~\eqref{eq:cluster_diameter} due to Lemma~\ref{lem:bounded_infnorm}. Let us condition on this event occurs from now on.

Algorithm~\ref{alg:cluster} constructs a set of disjoint $L_\infty$-balls of radius $2\gamma$, of which each is centered at one sample in $T$ and contains at least an $\alpha$-fraction of samples in $T$. Therefore, the number of such balls is at most $m = \lFloor1/\alpha\rFloor$. Denote the set by $\{\mathbb{B}_1,\dots,\mathbb{B}_m\}$. Let $\mathbb{B}_i'$ be the ball that has the same center as $\mathbb{B}_i$ but with $\ell_\infty$-radius of $6\gamma$. In the following, we show that there exists $ i\in[m]$, such that $T_i=T\cap\mathbb{B}_i'$ is $\alpha$-good.

Consider a sample $x\in \Sgood$, for which we know that $\infnorm{x-\mu}\leq \gamma$. Then, for the $L_\infty$-ball $\mathbb{B}_x := \{y\in\R^d: \infnorm{y-x} \leq 2\gamma \}$, all of the samples in $\Sgood$ will be contained in $\mathbb{B}_x$. In addition, there must exist one $\mathbb{B}_i$ that intersects $\mathbb{B}_x$, as otherwise $\mathbb{B}_x$ will be in the set $\{\mathbb{B}_1,\dots,\mathbb{B}_m\}$. That is, $\exists z\in T, z\in\mathbb{B}_x\cap\mathbb{B}_i$. By construction, $\mathbb{B}_x$ must be containted in $\mathbb{B}_i'$. Therefore, all samples of $\Sgood$ must be included in $T_i$ and $T_i$ is $\alpha$-good.
\end{proof}


\subsection{Proof of Lemma~\ref{lem:function_value_bound}}
\begin{proof}
Recall that after running Algorithm~\ref{alg:cluster}, every subset $T_i$ is contained in an $L_\infty$-ball of radius $6\gamma$. By Jensen's inequality and the convexity of the $L_{\infty}$-norm, we have  for all $x \in T$, $\infnorm{x- {\mu_T}}\leq 6\gamma$.


Recall that we assumed $p(x) = h_A(x - \mu_T)$. Thus $\E_{x\sim\N({\mu_T},\I)}[p(x)]=0$ due to the definition of harmonic polynomials. Thus, $\Var_{x\sim\N(\mu_T,\I)}[p(x)]=\twonorm{A}^2$. Denote $z=x-\mu_T$. Then,
\begin{align}\label{eq:tmp:mehler}
\abs{p(x)} = \abs{\sum_{j\in[k^{2\ell}]} c_{\ba^{(j)}}\frac{\hermite_{\ba^{(j)}}(z)}{\sqrt{\onenorm{\ba^{(j)}}!}} } 
\leq \sqrt{\(\sum_{j\in[k^{2\ell}]}c_{\ba^{(j)}}^2\)
\(\sum_{j\in[k^{2\ell}]}\frac{\hermite_{\ba^{(j)}}(z)^2}{\onenorm{\ba^{(j)}}!}\)}.
\end{align}
where $\ba^{(j)}$ is a $d$-dimensional multi-index for the $j$-th monomial, and $c_{\ba^{(j)}}$ denotes its coefficient. Observe that in the first step, $p(x)$ is written as a linear combination of $k^{2\ell}$ Hermite polynomials, since we are considering $p \in \mathbb{P}(\R^d, l, k^{2\ell}, 2\ell k^{2\ell})$. Note also that $\sum_{j\in[k^{2\ell}]}c_{\ba^{(j)}}^2 = \twonorm{A}^2 \leq 1$. 


To bound the second factor on the right-hand side of \eqref{eq:tmp:mehler}, we use Mehler's formula, which shows that for any $u$ with $\abs{u}<1$ and any natural number $a$,
\begin{equation*}
\sum_{a=0}^{\infty} \frac{\hermite_a^2(z_i)u^a}{a!} = \frac{1}{\sqrt{1-u^2}} e^{\frac{u}{1+u}{z}_i^2},
\end{equation*}
Since each $\hermite_{\ba^{(j)}}(z)$ has degree at most $l$, it can be decomposed as a product of at most $l$ univariate Hermite polynomials. Thus, we take such product and sum over $j \in [k^{2\ell}]$ to obtain
\begin{align*}
\sum_{j\in[k^{2\ell}]} \frac{ \prod_{\ba^{(j)}_i\neq0} \Big(\hermite_{\ba^{(j)}_i}(z_i)^2 \cdot u^{\ba^{(j)}_i} \Big) }{\onenorm{\ba^{(j)}}!}  
\leq k^{2\ell} \cdot (1-u^2)^{-\frac{{l}}{2}} \cdot  e^{\frac{u}{1+u}\twonorm{\hard_l(z)}^2}.
\end{align*}
To simplify the above expression, observe that $\prod_{\ba_i^{(j)} \neq 0} u^{\ba_i^{(j)}} = u^{\onenorm{\ba^{(j)}}} \geq u^l$. In addition, $\twonorm{\hard_l(z)}^2 \leq l \cdot \infnorm{z}^2 \leq 36 l \gamma^2$. Lastly, by algebra, $(1 - u^2)^{-\frac{l}{2}} \leq e^{\frac{u^2 l}{2}}$. Putting all pieces together gives
\begin{equation*}
\sum_{j\in[k^{2\ell}]}\frac{\hermite_{\ba^{(j)}}(z)^2}{\onenorm{\ba^{(j)}}!} \leq u^{-l} \cdot k^{2\ell} \cdot e^{\frac{u^2 l}{2}} \cdot e^{\frac{36l \gamma^2 u}{1+u}} = k^{2\ell} \cdot u^{-l} \cdot e^{\frac{u^2 l}{2} + \frac{36l \gamma^2 u}{1+u}}.
\end{equation*}
We set $u=\frac{1}{\gamma}$; this is possible as $\gamma > 1$. Then the exponent $\frac{u^2 l}{2} + \frac{36l \gamma^2 u}{1+u} = \frac{l}{2\gamma^2} + \frac{36l}{1+1/\gamma^2} \leq 37l$. Without loss of generality, we may assume that $\gamma > e^{37}$; in fact, we can always ensure this by setting $\gamma = (C_0 +e^{37}) \cdot \sqrt{\ell \cdot \log\frac{\ell d}{\alpha \tau}}$ where $C_0$ is the constant given in Algorithm~\ref{alg:cluster}. Thus, it follows that
\begin{equation*}
\sum_{j\in[k^{2\ell}]}\frac{\hermite_{\ba^{(j)}}(z)^2}{\onenorm{\ba^{(j)}}!} \leq  k^{2\ell} \cdot \gamma^l \cdot e^{37l} \leq k^{2\ell} \cdot \gamma^{2l}.
\end{equation*}
Plugging it into \eqref{eq:tmp:mehler} completes the proof.
\end{proof}

\section{Analysis of \aefilter}\label{sec:app:proof-main}


We collect a few useful facts about Hermite polynomials.

Recall that for an order-$l$ tensor $A\in\R^{d^l}$, $\twonorm{A}$ denotes its $L_2$ norm by seeing it as a long vector, and for a polynomial $p:\R^d\rightarrow\R$, $\twonorm{p} := \E_{x \sim N(0,\I_d)}[p^2(x)])^{1/2}$. 

The following can be easily seen from the definition of harmonic polynomials.
\begin{fact}\label{fact:two_norms}
For all order-$l$ symmetric tensors $A$ and its corresponding harmonic polynomial $h_A$, we have that $\twonorm{h_A} = \twonorm{A}$. Moreover, if $l>0$, then $\E_{x\sim N(0,\I_d)}[h_A(x)]=0$.
\end{fact}

\begin{claim}\label{cl:uni-to-multivariate}
Let $v\in\R^d$ be a unit vector. For $x\in\R^d$, the polynomial $p(x)=\hermite_{l}(v\cdot x)$  is harmonic with respect to $x$ with degree $l$. That is, there exists a tensor $A = \tensor(p)$ which is symmetric and with order $l$.
\end{claim}

%


\subsection{Proof of Lemma~\ref{lem:returned_vector}}

\begin{proof}

Recall that we denoted $\lambdasparse = C_1\cdot\big[(\ell+ C_1 \log\frac{1}{\alpha}) \cdot\log^2(2+\log\frac{1}{\alpha}) \big]^{2\ell}$ in Algorithm~\ref{alg:mainSub-highdegree}.

Observe that if $\fronorm{\tilde{\Sigma}_{U}} \leq \lambdasparse$, 
then for any index set $\Omega \subset [d^{\ell}]$ with $\abs{\Omega} \leq k^{\ell}$, we have
\begin{equation*}
\lambda_{\max}({\tilde{\Sigma}_{\Omega\times\Omega}}) \leq \fronorm{\tilde{\Sigma}_{\Omega\times\Omega}} \leq \fronorm{\tilde{\Sigma}_{U}} \leq \lambdasparse,
\end{equation*}
where $\lambda_{\max}(\cdot)$ denotes the maximum eigenvalue and the second step follows from our choice of $U$ which maximizes the restricted Frobenius norm.

Thus, for any $u\in\R^{d^{\ell}}$ with $\zeronorm{u}\leq k^{\ell}$,
\begin{equation}
u^{\top}\tilde{\Sigma}u \leq \lambda_{\max}({\tilde{\Sigma}_{\Omega\times\Omega}}) \leq \lambdasparse.
\end{equation}

Let $v$ be a  $k$-sparse unit vector in $\R^d$. That is, $v\in\R^d,\zeronorm{v}\leq k,\twonorm{v}=1$. Consider some symmetric order-$\ell$ tensor $B$ such that $\hermite_{\ell}(v\cdot(x-\mu_T)) = h_{B}(x-\mu_T)$ (Claim~\ref{cl:uni-to-multivariate}). Due to the sparsity of $v$, we know that $B$ is an outer product of $\ell$ number of $k$-sparse vectors; hence $\zeronorm{B} \leq k^{\ell}$. As $h_B(x-\mu_T)$ is a degree-$\ell$ harmonic polynomial and the vector $P_{d,\ell}(x-\mu_T)$ includes all Hermite polynomials with degree exactly $\ell$, we know that we can write $h_B(x - \mu_T) = u_B \cdot P_{d,\ell}(x- \mu_T)$ for some $u_B\in\R^{d^\ell},\zeronorm{u_B}\leq k^\ell$. 
Thus, we have that
\begin{equation*}
\E[h_B(T-\mu_T)^2] = \E[(u_B \cdot P_{d,\ell}(T-\mu_T))^2] = u_B^{\top}\tilde{\Sigma}u_B \leq \lambdasparse\twonorm{u_B}^2 = \lambdasparse \twonorm{B}^2.
\end{equation*}

By Fact~\ref{fact:two_norms}, observe that $\twonorm{B}^2 = \E_{x \sim \N(\mu_T,\I_d)}[h_B(x-\mu_T)^2] = \ell!$, and thus we have $\E[\hermite_\ell(v\cdot(T-\mu_T))^2] = \E[h_B(T-\mu_T)^2] \leq \lambdasparse\ell!$. 

As a result, we have for any $k$-sparse unit vector $v\in\R^d$ that
\begin{align}
\E[\hermite_\ell(v\cdot(\Sgood\cap T-\mu_T))^2] 
&= \frac{1}{\abs{\Sgood\cap T}}\sum_{x\in \Sgood\cap T}\hermite_\ell(v\cdot(x-\mu_T))^2 \notag \\ 
&\leq \frac{1}{\alpha\cdot \abs{T}} \sum_{x\in T}\hermite_\ell(v\cdot(x-\mu_T))^2 \notag \\ 
&= \frac{1}{\alpha}\cdot\E[\hermite_\ell(v\cdot(T-\mu_T))^2]
\leq \frac{\lambda^*_{\text{sparse}}\cdot \ell!}{\alpha}, \label{eq:sparse-univariate-variance}
\end{align}
where the first inequality follows from the condition that $T$ is $\alpha$-good, which, by Definition~\ref{def:good-set}, implies $\abs{\Sgood \cap T} / \abs{T} \geq \alpha$.


The remaining analysis borrows the proof strategy from ~\cite{dia2018list}. In particular,
we will need the following lemma.
\begin{lemma}[Lemma~3.34 of ~\cite{dia2018list}]
For any $v\in\R^d$, the polynomial $\hermite_{l}(v\cdot(G-\mu_T))$ has mean $(v\cdot(\mu-\mu_T))^{l}$ and variance at most $2\max(l,v\cdot(\mu-\mu_T))^{2(l-1)}$.
\end{lemma}

Now to ease the notation, write $\theta := v\cdot(\mu-\mu_T)$. By Cantelli's inequality we have 
\begin{align*}
&\Pr\[\hermite_\ell(v\cdot(G-\mu_T)) \geq \theta^{\ell}-\sqrt{2}\max(\ell,\theta)^{(\ell-1)} \] \\
&\geq 1-\frac{\Var[\hermite_\ell(v\cdot(G-\mu_T))]}{\Var[\hermite_\ell(v\cdot(G-\mu_T))]+\Var[\hermite_\ell(v\cdot(G-\mu_T))]} \geq 1-\frac12 = \frac12.
\end{align*}

Since $\Sgood$ is representative, by Definition~\ref{def:representative-set}
\begin{equation*}
\Pr\[\hermite_\ell(v\cdot(\Sgood-\mu_T)) \geq \theta^{\ell} - \sqrt{2}\max(\ell,\theta)^{(\ell-1)} \] \geq \frac12-\frac{\alpha^3}{100} \geq \frac{49}{100}.
\end{equation*}
Since $T$ is $\alpha$-good, due to Definition~\ref{def:good-set}, $\abs{\Sgood\cap T}/\abs{\Sgood} \geq 1-\frac{\alpha}{6}\geq\frac{80}{100}$, we have that
\begin{equation*}
\Pr\[\hermite_\ell(v\cdot(\Sgood\cap T-\mu_T)) \geq \theta^{\ell} - \sqrt{2}\max(\ell,\theta)^{(\ell-1)} \] \geq \frac{49}{100} - \frac{20}{100} \geq \frac14.
\label{eq:geq14}
\end{equation*}

On the other hand, due to Eq.~\eqref{eq:sparse-univariate-variance}, applying Markov's inequality gives that for any $k$-sparse unit vector $v$,
\begin{align}
\Pr\[\hermite_\ell(v\cdot(\Sgood\cap T-\mu_T)) \geq \sqrt{\frac{4\lambda^*_{\text{sparse}}\cdot \ell!}{\alpha}} \] &\leq \frac{\E[\hermite_\ell(v\cdot(\Sgood\cap T-\mu_T))^2]}{\(\sqrt{\frac{4\lambda^*_{\text{sparse}}\cdot \ell!}{\alpha}}\)^2} \notag \\
&\leq \frac{\lambda^*_{\text{sparse}}\cdot \ell!/\alpha}{4\lambda^*_{\text{sparse}}\cdot \ell!/\alpha} = \frac14.
\label{eq:leq14}
\end{align}

Recall that $\theta = v\cdot(\mu-\mu_T)$. From Eq. \eqref{eq:geq14} and \eqref{eq:leq14}, we have that for any $k$-sparse unit vector $v\in\R^d$,
\begin{equation*}
(v\cdot(\mu-\mu_T))^{\ell} - \sqrt{2}\max(\ell,v\cdot(\mu-\mu_T))^{(\ell-1)} \leq \sqrt{\frac{4\lambda^*_{\text{sparse}}\cdot \ell!}{\alpha}}.
\end{equation*}
Note that $\theta^{\ell} \leq \sqrt{2}\max(\ell,\theta)^{(\ell-1)}$ only when $\theta \leq 2\ell$, and so we have that for any $k$-sparse unit vector $v\in\R^d$,
\begin{align*}
v\cdot(\mu-\mu_T) &\leq 2\ell + \Big(\frac{4\lambda^*_{\text{sparse}}\cdot \ell!}{\alpha}\Big)^{\frac{1}{2\ell}} \\
&= O\Big(2\ell + \Big(\frac{4(C_1\cdot\big[(\ell+ C_1 \log\frac{1}{\alpha}) \cdot\log^2(2+\log\frac{1}{\alpha}) \big]^{2\ell})\cdot \ell!}{\alpha}\Big)^{\frac{1}{2\ell}}
\Big) \\
&= O\Big( \alpha^{-\frac{1}{2\ell}} \cdot \sqrt{\ell}  \Big(\ell+\log\frac{1}{\alpha}\Big)\cdot\log^2\Big(2+\log\frac{1}{\alpha}\Big) \Big) .
\end{align*}
By choosing $v = \hard_k(\mu - \mu_T)$ and combining the above with  Lemma~\ref{lem:sparse_vectors}, we complete the proof.
\end{proof}

\subsection{Analysis of \bmf}


Recall the notations in \bmf (Algorithm~\ref{alg:basic}): $R={(C_1\cdot\log(\frac{1}{\alpha}))}^{l/2}$, $\gamma=C_0 \cdot\sqrt{\ell \cdot \log(\frac{ \ell d}{\alpha \tau})}$, and the length of the interval $[a, b]$, i.e. $b-a$, equals $C_1\cdot R\cdot\log(2+\log\frac{1}{\alpha})$. We will need a series of results to prove Theorem~\ref{thm:bmf}. First, we note that if \bmf returns at Step~\ref{step:BMF-return-no}, then $T$ must not be $\alpha$-good in view of Lemma~\ref{lem:bounded_infnorm}. Thus we only need to consider the remaining steps. In particular, we divide the output of \bmf into three cases: 
\begin{itemize}
\item \textsc{Case} 1: it returns $\TBD$ at Step~\ref{step:BMF-return-yes}.

\item \textsc{Case} 2: it returns one subset $\{(T', \alpha')\}$ at Step~\ref{step:BMF-return-onecluster}.

\item \textsc{Case} 3: it returns two subsets $\{(T_1, \alpha_1), (T_2, \alpha_2)\}$ at Step~\ref{step:BMF-return-twocluster}.
\end{itemize}

We analyze the performance for each case in the following.

\subsubsection{Analysis of \textsc{Case} 1}

\begin{proposition}\label{prop:case-1}
Consider  Algorithm~\ref{alg:basic}.
If it returns $\TBD$ and if $T$ is an $\alpha$-good set, then {$\abs{\E[p(G)] - \E[p(T)]} \leq O\big( {(\ell +\log\frac{1}{\alpha})}^{\frac{l}{2}}\cdot\log(2+\log\frac{1}{\alpha})\big)$}.
\end{proposition}
\begin{proof}
We first argue that most of the good samples in $T$ have $p(x)$ value close to $\E[p(G)]$.
\begin{claim}\label{cl:good_samples_fraction}
If $T$ is $\alpha$-good, then the samples $x\in T\cap \Sgood$ that satisfy $\abs{p(x)-\E[p(G)]} < R$ constitute at least an $\(\alpha-\frac{\alpha^3}{100}\)$-fraction of $T$ and an $(1-\frac{\alpha}{6}-\frac{\alpha^3}{100})$-fraction of $\Sgood$.
\end{claim}

Next, we claim that if there exists an appropriate interval $[a,b]$ in Step~\ref{step:BMF-find-interval}, then the mean of $p(G)$ is in the interval $[a-R,b+R]$.
\begin{claim}\label{cl:location_EpG}
If $T$ is $\alpha$-good, and the interval $[a,b]$ contains at least $(1-\frac{\alpha}{2})$-fraction of values of $p(x)$ for $x\in T$, then $\E[p(G)] \in [a-R,b+R]$.
\end{claim}

Now by construction, if Algorithm~\ref{alg:basic} returns $\TBD$, then 
\begin{equation}
\Var[p(T)] \leq C_1 \cdot{\(\ell+C_1\log\frac{1}{\alpha}\)}^l \cdot \log^2 \(2+\log\frac{1}{\alpha} \).
\end{equation}
On the other hand, the interval $[a,b]$ contains at least $(1-\frac{\alpha}{2})$ fraction of values of $p(x)$ for $x\in T$. Therefore, the contribution of the samples in $[a,b]$ to the variance gives
\begin{equation}
\Var[p(T)] \geq \(1 - \frac{\alpha}{2} \) \cdot \max\Big\{0, \abs{\E[p(T)] - \frac{a+b}{2}} - \frac{b-a}{2}\Big\}^2.
\end{equation}
To see this, note that $\frac{a+b}{2}$ is the midpoint and $\frac{b-a}{2}$ is the length of interval $[a,b]$. When $\E[p(T)]$ is inside the interval, $\abs{\E[p(T)] - \frac{a+b}{2}} - \frac{b-a}{2}<0$ and the variance is lowered bounded by $0$. Otherwise, when $\E[p(T)]$ is outside the interval, the distance from any sample in $[a,b]$ to $\E[p(T)]$ is at least $\abs{\E[p(T)] - \frac{a+b}{2}} - \frac{b-a}{2}\geq0$. 

Moreover, since $b-a\leq O( (\log(1/\alpha))^{l/2}\cdot\log(2+\log(1/\alpha)))$,
\begin{equation}
\abs{\E[p(T)] - (a+b)/2} \leq \frac{b-a}{2} + \sqrt{\Var[p(T)]} = O({(\ell+C\log(1/\alpha))}^{l/2}\log(2+\log(1/\alpha))).
\end{equation}

From the Claim~\ref{cl:location_EpG}, we also have
\begin{equation}
\abs{\E[p(G) - (a+b)/2]} \leq \frac{b-a}{2} + R = O({(\ell+C\log(1/\alpha))}^{l/2}\log(2+\log(1/\alpha))).
\end{equation}
By the triangle inequality, we have that $\abs{\E[p(G)] - \E[p(T)]} = O({(\ell+C\log(1/\alpha))}^{l/2}\log(2+\log(1/\alpha)))$.
\end{proof}

\begin{proof}[Proof of Claim~\ref{cl:good_samples_fraction}]
Since $T$ is $\alpha$-good, and $\Var[p(G)]\leq 1$. By degree-$l$ Chernoff bound (Lemma~\ref{lem:chernoff}) and definition of representative set (Definition~\ref{def:representative-set}), for $R = (C_1 \cdot\log(1/\alpha))^{l/2}$
\begin{align*}
\Pr[\abs{p(\Sgood)-\E[p(G)]} \geq R] &\leq e^{-\Omega(R^{2/l})} + \frac{\alpha^3}{100{k^{2\ell}\cdot\log^l(\frac{\ell d}{\alpha\tau})}} \\
&\leq e^{-C\cdot\log(1/\alpha)} + \frac{\alpha^3}{100{k^{2\ell}\cdot\log^l(\frac{\ell d}{\alpha\tau})}} \\
&= \alpha^C + \frac{\alpha^3}{100{k^{2\ell}\cdot\log^l(\frac{\ell d}{\alpha\tau})}} \leq \frac{\alpha^3}{100},
\end{align*}
for large enough constant $C>0$.
\end{proof}

\begin{proof}[Proof of Claim~\ref{cl:location_EpG}]
From Claim~\ref{cl:good_samples_fraction}, at least an $(\alpha-\alpha^3/100)$-fraction of $T$ is $R$-close to $\E[p(G)]$. Also we know that at most an $\frac{\alpha}{2}$-fraction of $T$ are not in $[a,b]$ by the definition of the interval $[a, b]$. Then, there must be at least
\begin{equation*}
\(\alpha-\frac{\alpha^3}{100}\) -\frac{\alpha}{2} = \frac{\alpha}{2} - \frac{\alpha^3}{100} = \frac{\alpha}{2}\(1-\frac{\alpha^2}{50}\) > 0
\end{equation*}
fraction of samples in $T$ that are in $[a,b]$ and $R$ close to $\E[p(G)]$.
Therefore, $\E[p(G)]$ must be in $[a-R,b+R]$.
\end{proof}

\subsubsection{Analysis of \textsc{Case} 2}

\begin{lemma}\label{lem:threshold_for_one}
Consider  Algorithm~\ref{alg:basic}.
If it reaches Step~\ref{step:BMF-onecluster-threshold}, there must exist a threshold $t>2R$ satisfying the inequality thereof.
\end{lemma}
\begin{proof}
We will prove this lemma by contradiction. Assume that Algorithm~\ref{alg:basic} reaches Step~\ref{step:BMF-onecluster-threshold}, but for all $t>2R$, we have 
\begin{equation*}
\Pr[\min\{\abs{p(T)-a},\abs{p(T)-b}\} \geq t] \leq \frac{32}{\alpha}\exp(-(t-2R)^{2/l}) +  {\frac{2\alpha^2}{ k^{2\ell} \log^{l}(\frac{\ell d}{\alpha \tau})} }.
\end{equation*}
By change of variables, we have that for any $t>2R+\frac{b-a}{2}$,
\begin{equation*}
\Pr\[\abs{p(T)-\frac{a+b}{2}} \geq t\] \leq \frac{32}{\alpha}e^{-\(t-2R-\frac{b-a}{2}\)^{2/l}} + {\frac{2\alpha^2}{ k^{2\ell} \log^{l}(\frac{\ell d}{\alpha \tau})} }.
\end{equation*}
Note that this inequality only holds non-trivially when $t \geq t_0$ where $t_0 =2R + \frac{b-a}{2} + (\log\frac{32}{\alpha})^{l/2}$; namely, if $t < t_0$, the right-hand side is at least $1$.

By Lemma~\ref{lem:function_value_bound}, we have $\max_{x,y\in T} \abs{p(x) - p(y)} \leq 2 k^{\ell} \cdot \gamma^{l}$, where $\gamma=C_0 \cdot\sqrt{\ell \cdot \log(\frac{ \ell d}{\alpha \tau})}$. Also note that the size of the interval $[a, b]$ equals $C_1 \cdot R \cdot \log(2 + \log\frac{1}{\alpha})$ which is less than $k^{\ell} \cdot \gamma^{l}$. Therefore, 
\begin{equation}\label{eq:tmp:max-p}
\max_{x\in T} \abs{p(x) - \frac{a+b}{2}} \leq 3 k^{\ell} \cdot \gamma^{l}.
\end{equation}

Then, we have that
\begin{align*}
\Var[p(T)] &\leq \E\[\(p(T)-\frac{a+b}{2}\)^2\] \\
&= \int_{0}^{\infty} \Pr \Big[ \big(p(T)-\frac{a+b}{2}\big)^2 \geq t^2 \Big] d t^2 \\
&\stackrel{\zeta_1}{=} 2 \int_{0}^{{3k^{\ell}\cdot\gamma^{l}} }\Pr\[\abs{p(T)-\frac{a+b}{2}}\geq t\]tdt \\
&= 2 \int_{0}^{t_0} \Pr\[\abs{p(T)-\frac{a+b}{2}}\geq t\]tdt  + 2 \int_{t_0}^{3k^{\ell}\cdot\gamma^{l}} \Pr\[\abs{p(T)-\frac{a+b}{2}}\geq t\]tdt \\
&\leq t_0^2 + 2 \int_{t_0}^{{3k^{\ell}\cdot\gamma^{l}} } \Big( \frac{32}{\alpha}e^{-\(t-2R-\frac{b-a}{2}\)^{2/l}} + {\frac{2\alpha^2}{k^{2\ell}\cdot\log^{l}(\frac{\ell d}{\alpha\tau}) } } \Big)  tdt \\
&= t_0^2 + {\frac{2\alpha^2}{k^{2\ell}\cdot\log^{l}(\frac{\ell d}{\alpha\tau}) } } \cdot 9k^{2\ell} \cdot \gamma^{2l}  +\frac{32}{\alpha}\int_{(\log\frac{32}{\alpha})^{{l}/{2}}}^{\infty}e^{-t^{2/l}}\cdot (2t+4R+b-a)dt \\
&= t_0^2 + 18 C_0^2 \cdot \alpha^2 \cdot \ell^l + \frac{32}{\alpha}\int_{\log\frac{32}{\alpha}}^{\infty}e^{-u}\cdot (2u^{l/2}+4R+b-a)\cdot \frac{l}{2}\cdot u^{\frac{l}{2}-1}du\\
&\stackrel{\zeta_2}{\leq} O\(\(\log\({1}/{\alpha}\)\)^{l}\cdot\log^2(2+\log(1/\alpha))\) + O\({2\alpha^2} \cdot {\ell^{l}} \) \\  &\quad\quad\quad + O((l+\log{1}/{\alpha})^{l}\cdot\log(2+\log{1}/{\alpha}))\\
&\leq O\( {\(\ell + \log\({1}/{\alpha}\)\)}^{l} \cdot\log^2(2+\log(1/\alpha))\) ,
\end{align*}
where $\zeta_1$ holds in view of \eqref{eq:tmp:max-p}, and
where $\zeta_2$ follows since
\begin{align*}
&\quad\frac{32}{\alpha}\int_{\log(\frac{32}{\alpha})}^{\infty}e^{-u}\cdot (2u^{\frac{l}{2}}+4R+b-a)\cdot\frac{l}{2}\cdot u^{\frac{l}{2}-1}du \\
&= \frac{32}{\alpha}\int_{\log(\frac{32}{\alpha})}^{\infty}e^{-u}\cdot 2u^{l-1}\cdot\frac{l}{2}du + \frac{32}{\alpha}\int_{\log(\frac{32}{\alpha})}^{\infty}e^{-u}(4R+b-a)\cdot\frac{l}{2}\cdot u^{\frac{l}{2}-1}du \\
&= \frac{32}{\alpha}\cdot2\cdot\frac{l}{2}\int_{\log(\frac{32}{\alpha})}^{\infty}e^{-u}\cdot u^{l-1}du + \frac{32}{\alpha}\cdot(4R+b-a)\cdot\frac{l}{2}\int_{\log(\frac{32}{\alpha})}^{\infty}e^{-u}\cdot u^{\frac{l}{2}-1}du \\
&\stackrel{\zeta_3}{\leq} \frac{32}{\alpha}\cdot2\cdot\frac{l}{2}\cdot e^{-\log\frac{32}{\alpha}}\cdot\Big(\log\frac{32}{\alpha}+l\Big)^{l-1} + \frac{32}{\alpha}\cdot(4R+b-a)\cdot\frac{l}{2}\cdot e^{-\log\frac{32}{\alpha}}\cdot\Big(\log\frac{32}{\alpha}+\frac{l}{2}\Big)^{\frac{l}{2}-1}  \\
&\leq \Big(\log\frac{32}{\alpha}+l\Big)^{l} + (4R+b-a)\cdot\Big(\log\frac{32}{\alpha}+\frac{l}{2}\Big)^{\frac{l}{2}}  \\
&\leq \Big(\log\frac{32}{\alpha}+l\Big)^{l} + \Big(4{\big(C_1\cdot\log\frac{1}{\alpha}\big)}^{l/2}+C_1\cdot R\cdot\log\big(2+\log\frac{1}{\alpha}\big)\Big)\cdot\Big(\log\frac{32}{\alpha}+\frac{l}{2}\Big)^{\frac{l}{2}}  \\
&= O\Big(\Big(l+\log\frac{1}{\alpha}\Big)^{l}\cdot\log\Big(2+\log\frac{1}{\alpha}\Big)\Big),
\end{align*}
where $\zeta_3$ is due to the incomplete gamma function (see Claim~3.11 of~\cite{dia2018list}), i.e. $\int_{x}^{\infty}e^{-t}\cdot t^{s-1}dt \leq e^{-x}(x+s)^{s-1}$, for $s\geq1,x\geq0$.

In other words, had we not found an appropriate threshold $t>2R$ at Step~\ref{step:BMF-onecluster-threshold}, Algorithm~\ref{alg:basic} would have returned at Step~\ref{step:BMF-return-yes}, which is a contradiction. 
This completes the proof.
\end{proof}

Once we have verified the existence of such threshold $t$, it is easy to see that the resultant $T'$ is a subset of $T$, and $\alpha' \geq \alpha$ by algebraic calculation. This has been already shown in \cite{dia2018list}.

\begin{lemma}[Lemma~3.13 of~\cite{dia2018list}]\label{lem:alpha'>alpha}
Consider Algorithm~\ref{alg:basic}. If it reaches Step~\ref{step:BMF-return-onecluster}, then the output $\{(T', \alpha')\}$ is such that $T' \subset T$ and $\alpha'>\alpha$.
\end{lemma}

Next, we show that \bmf sanitizes the sample set, i.e. it removes more corrupted samples than the uncorrupted ones.
\begin{lemma}\label{lem:T'-good}
Consider Algorithm~\ref{alg:basic}. If it reaches Step~\ref{step:BMF-return-onecluster}, and if $T$ is $\alpha$-good and $\Var[p(G)] \leq 1$, then the output $\{(T', \alpha')\}$ is such that $T'$ is $\alpha'$-good.
\end{lemma}
\begin{proof}
Due to Algorithm~\ref{alg:cluster}, the $\ell_\infty$-distance among all pairs of the samples are bounded. It remains to show $\abs{\Sgood\cap T'}/\abs{T'} \geq \alpha'$ and $\abs{\Sgood\cap T'}/\abs{\Sgood} \geq 1-\alpha'/6$. 

We claim that for any $t>2R$, the following holds:
\begin{equation}\label{eq:tmp:min-p(S)}
\Pr[\min\{\abs{p(\Sgood)-a},\abs{p(\Sgood)-b}\} \geq t] \leq 2e^{-(t-R)^{2/l}} + \frac{\alpha^3}{50{k^{2\ell}\cdot\log^l(\frac{\ell d}{\alpha\tau})} }.
\end{equation}
To see the rationale, we note that by Claim~\ref{cl:location_EpG}, we have $\E[p(G)] \in[a-R,b+R]$.  Since $\E[p(G)]-R\leq b$, we have 
\begin{align*}
\Pr[p(\Sgood)-b\geq t] &\leq \Pr[p(\Sgood)-(\E[p(G)]-R)\geq t] \\
&= \Pr[p(\Sgood)- \E[p(G)]\geq t-R] \\
&\leq \Pr[p(G)-E[p(G)]\geq t-R] + \frac{\alpha^3}{100{k^{2\ell}\cdot\log^l(\frac{\ell d}{\alpha\tau})} }\\
&\leq e^{-(t-R)^{2/l}} + \frac{\alpha^3}{100{k^{2\ell}\cdot\log^l(\frac{\ell d}{\alpha\tau})} },
\end{align*}
where in the third step, we used the fact that $\Sgood$ is representative (see Definition~\ref{def:representative-set}), and in the last step we applied Lemma~\ref{lem:chernoff}.


The inequality \eqref{eq:tmp:min-p(S)} follows since $\min\{ \abs{p(\Sgood) -a}, \abs{p(\Sgood)-b} \} \geq t$ is a subevent of $\abs{p(\Sgood)-b}>t$. 

Since $T$ is $\alpha$-good, we know that a  $1-\frac{\alpha}{6}\geq\frac12$ fraction of the samples in $\Sgood$ is in $\Sgood\cap T$. Therefore,
\begin{equation}
\Pr[\min\{\abs{p(\Sgood\cap T)-a},\abs{p(\Sgood\cap T)-b}\} \geq t] \leq 4e^{-(t-R)^{2/l}} + \frac{\alpha^3}{25{k^{2\ell}\cdot\log^l(\frac{\ell d}{\alpha\tau})} }.
\end{equation}
Due to the inequality of Step~\ref{step:BMF-onecluster-threshold} in Algorithm~\ref{alg:basic}, we know that the above probability is at least $8/\alpha$ times larger for the samples in $T$. 
Therefore,
\begin{align*}
\frac{\abs{\Sgood\cap T'}}{\abs{T'}} &=  \frac{\abs{\Sgood\cap T'}}{\abs{\Sgood\cap T}} \frac{\abs{\Sgood\cap T}}{\abs{T}}\frac{\abs{T}}{\abs{T'}} \\
&\geq \(1-\frac{\alpha}{8}\cdot\(1-\frac{\abs{T'}}{\abs{T}}\)\) \cdot \alpha \cdot\frac{\abs{T}}{\abs{T'}}\\
&\geq \( \(1-\frac{\alpha}{8}\)\cdot \frac{\abs{T}}{\abs{T'}} + \frac{\alpha}{8} \) \cdot\alpha \\
&= \alpha' ,
\end{align*}
meaning that the remaining fraction of good samples in $T'$ is at least $\alpha'$.

On the other hand, since ${\abs{\Sgood\cap T}}/{\abs{\Sgood}} \geq 1-\alpha/6$ and $\Big(1-\frac{\alpha}{8}\cdot\Big(1-\frac{\abs{T'}}{\abs{T}}\Big)\Big)\alpha = \alpha'\abs{T'}/\abs{T}$, we have
\begin{align*}
\frac{\abs{\Sgood\cap T'}}{\abs{\Sgood}} &= \frac{\abs{\Sgood\cap T'}}{\abs{\Sgood\cap T}} \frac{\abs{\Sgood\cap T}}{\abs{\Sgood}} \\
&\geq \(1-\frac{\alpha}{8}\cdot\(1-\frac{\abs{T'}}{\abs{T}}\)\) \(1-\frac{\alpha}{6}\) \\
&= \(1-\frac{\alpha}{8}\cdot\(1-\frac{\abs{T'}}{\abs{T}}\)\) - \frac{\alpha'\abs{T'}}{6\abs{T}} ,
\end{align*}
thus,
\begin{align*}
\frac{\abs{\Sgood\cap T'}}{\abs{\Sgood}} - \(1-\frac{\alpha'}{6}\) &\geq 1-\frac{\alpha}{8}\(1-\frac{\abs{T'}}{T}\) -\frac{\alpha'}{6}\frac{\abs{T'}}{\abs{T}} - \(1-\frac{\alpha'}{6}\) \\
&= \(\frac{\alpha'}{6}-\frac{\alpha}{8}\)\(1-\frac{\abs{T'}}{T}\) > 0
\end{align*}
This proves that $T'$ is $\alpha'$-good.
\end{proof}

We summarize the performance of \bmf in \textsc{Case}~2 in the following proposition, which is an immediate combination of Lemma~\ref{lem:threshold_for_one}, Lemma~\ref{lem:alpha'>alpha}, and Lemma~\ref{lem:T'-good}.

\begin{proposition}\label{prop:case-2}
Consider Algorithm~\ref{alg:basic}. If it reaches Step~\ref{step:BMF-onecluster-threshold}, there must exist $t > 2R$ that satisfies the inequality of this step, and the algorithm will output $\{(T', \alpha')\}$ with $T' \subset T$ and $\alpha' \geq \alpha$. If, in addition, $T$ is $\alpha$-good and $\Var[p(G)] \leq 1$, then $T'$ is $\alpha'$-good.
\end{proposition}

\subsubsection{Analysis of \textsc{Case} 3}

\begin{lemma}[Lemma~3.12 of~\cite{dia2018list}]\label{lem:threshold_for_two}
Consider  Algorithm~\ref{alg:basic}.
If it reaches Step~\ref{step:BMF-twocluster-threshold}, there must exist a threshold $t$ that satisfy the conditions thereof.
\end{lemma}

\begin{lemma}[Lemma~3.14 of~\cite{dia2018list}]\label{lem:two-mf-condition}
Consider  Algorithm~\ref{alg:basic}.
If it reaches Step~\ref{step:BMF-twocluster-threshold}, then the output $\{(T_1,\alpha_1),(T_2,\alpha_2)\}$ is such that $T_1 \subset T$, $T_2 \subset T$, and $\frac{1}{\alpha_1^2}+\frac{1}{\alpha_2^2} \leq \frac{1}{\alpha^2}$.
\end{lemma}


\begin{lemma}\label{lem:Ti-good}
Consider  Algorithm~\ref{alg:basic}.
If it reaches Step~\ref{step:BMF-twocluster-threshold}, and if $T$ is $\alpha$-good, then the output $\{(T_1,\alpha_1),(T_2,\alpha_2)\}$ is such that $T_i$ is $\alpha_i$-good for some $i \in \{1, 2\}$.
\end{lemma}
\begin{proof}
Recall that Claim~\ref{cl:good_samples_fraction} lower bounds the fraction of the good samples (i.e. $x\in \Sgood\cap T$) that satisfy $\abs{p(x)-\E[p(G)]}<R$. Since $T_1$ and $T_2$ overlap in an interval of length at least $2R$, the good samples must be contained in either one of both two clusters. We will show that the $T_i$ with these good samples (in interval of length $2R$) is $\alpha_i$-good.

Since $T$ is $\alpha$-good, we have $\abs{\Sgood\cap T}/\abs{T} \geq \alpha$ and $\abs{\Sgood\cap T}/\abs{\Sgood} \geq (1-\alpha/6)$. We want to show that (i) $\abs{\Sgood\cap T_i}/\abs{T_i} \geq \alpha_i$ and (ii) $\abs{\Sgood\cap T_i}/\abs{\Sgood} \geq (1-\alpha_i/6)$.

To show (i), note that $\abs{\Sgood\cap T_i} \geq \(\alpha-\frac{\alpha^3}{100}\) \abs{T}$ due to Claim~\ref{cl:good_samples_fraction}. Thus,
\begin{align*}
\frac{\abs{\Sgood\cap T_i}}{\abs{T_i}} = \frac{\abs{\Sgood\cap T_i}}{\abs{T}}\cdot\frac{\abs{T}}{\abs{T_i}} \geq \(\alpha-\frac{\alpha^3}{100}\)\cdot\frac{\abs{T}}{\abs{T_i}} =\alpha_i ,
\end{align*}
where the last transition is by definition.

To show (ii), we only have to show that $\alpha_i/6\geq\alpha/6+\alpha^3/100$, i.e. $\alpha_i\geq\alpha+3\alpha^3/50$. Note that $\abs{T}-\abs{T_i} \geq \frac{\alpha}{4}\abs{T}, \forall i$. Thus, $\abs{T}/\abs{T_i}\geq\frac{1}{1-\alpha/4}$ and we can show that
\begin{align*}
\alpha_i
\geq \alpha \cdot \frac{1-{\alpha^2}/{100}}{1-\alpha/4}  \geq \alpha \cdot \frac{100-\alpha}{100-25\alpha} \geq \alpha\(1+\frac{24\alpha}{100-25\alpha}\) \geq \alpha\(1+\frac{3\alpha^3}{50}\).
\end{align*}
This completes the proof.
\end{proof}

Combining Lemma~\ref{lem:threshold_for_two}, Lemma~\ref{lem:two-mf-condition}, and Lemma~\ref{lem:Ti-good}, we immediately have the following.

\begin{proposition}\label{prop:case-3}
Consider  Algorithm~\ref{alg:basic}.
If it reaches Step~\ref{step:BMF-twocluster-threshold}, then there must exist a threshold $t$ that satisfies the conditions in this step. Moreover, the output $\{(T_1, \alpha_1), (T_2, \alpha_2)\}$ is such that $T_1 \subset T$, $T_2 \subset T$, and $\frac{1}{\alpha_1^2} + \frac{1}{\alpha_2^2} \leq \frac{1}{\alpha^2}$. If, in addition, $T$ is $\alpha$-good, then $T_i$ is $\alpha_i$-good for some $i \in \{1, 2\}$.
\end{proposition}

\subsubsection{Proof of Theorem~\ref{thm:bmf}}

\begin{proof}
Observe that now Theorem~\ref{thm:bmf} is an immediate result by combining Proposition~\ref{prop:case-1}, Proposition~\ref{prop:case-2}, and Proposition~\ref{prop:case-3}.
\end{proof}

\subsection{Analysis of \harmf}\label{subsec:app:proof-hmf}

\subsubsection{Certifying the varaince of $p$ on $G$}

\begin{proof}[Proof of Lemma~\ref{lem:variance_G}]
The proof follows directly from Lemma~3.31 of \cite{dia2018list}. 
\end{proof}

\subsubsection{Analysis for $p_1$}

\begin{proof}[Proof of Lemma~\ref{lem:linear_filter}]
	First, if at any subroutine of \harmf, it returns ``$\NO$'' or a list of pairs $\{(T_i,\alpha_i)\}$, $\ANS \neq \TBD$.
	If that is not the case, it means \harmf reaches Step~\ref{step:HMF-certified} and \bmf returns ``$\TBD$'', then we have
	\begin{equation*}
	\Var[p_1(T)/\beta] \leq C_1\cdot {\Big(\ell+C_1\log\frac{1}{\alpha}\Big)}^{\ell} \cdot \log^2\Big(2+\log\frac{1}{\alpha}\Big),
	\end{equation*}
	because $p_1$ is of degree $\ell$. However, recall that the condition of Step~\ref{step:MF-linear-condition} in Algorithm~\ref{alg:mainSub-highdegree} is satisfied, thus
	\begin{align*}
	\Var[p_1(T)] &= \Var[v^*\cdot P_{d,\ell}(T - \mu_T)] \geq \lambda^* \geq \lambdasparse \\
	&={\Big[C_1\cdot\Big(\ell+ C_1 \log\frac{1}{\alpha}\Big) \cdot\log^2\Big(2+\log\frac{1}{\alpha}\Big) \Big]^{2\ell} } \\
	&\geq \Big(C_1\cdot\Big(1+\log\frac{1}{\alpha}\Big)\cdot\log^2\Big(2+\log\frac{1}{\alpha}\Big)\Big)^{\ell} \cdot C_1\cdot {\Big(\ell+C_1\log\frac{1}{\alpha}\Big)}^{\ell} \cdot \log^2\Big(2+\log\frac{1}{\alpha}\Big)\\
	&=\beta^2 \cdot C_1\cdot {\Big(\ell+C_1\log\frac{1}{\alpha}\Big)}^{\ell} \cdot \log^2\Big(2+\log\frac{1}{\alpha}\Big),
	\end{align*}
	which induces a contradition. We conclude that \bmf will not return ``$\TBD$'' at Step~\ref{step:HMF-certified}, which completes the proof.
\end{proof}


\subsubsection{Analysis for $p_2$}

\begin{proof}[Proof of Lemma~\ref{lem:quadratic_filter}]
	We see that the lemma holds as long as \harmf returns either ``$\NO$'' or a list of $(T_i,\alpha_i)$ for $p_2$ correctly. 
%
First, we claim that $p_2(x)$ is harmonic such that \mmf multifilters correctly at Step~\ref{step:HMF-callMMF}. To prove the claim, simply note that $p_2$ is of degree $2\ell$ and consists of a set of $k^{2\ell}$ Hermite polynomials. In addition, $p_2(x)$ only applies on a set of $2\ell k^{2\ell}$ coordinates.

Based on the correctness of \mmf, it remains to show that if every subroutine of \harmf returns ``$\TBD$'', then $T$ must not be $\alpha$-good. 
Consider that Algorithm~\ref{alg:hmf} reaches Step~\ref{step:HMF-certified}, and \bmf returns ``$\TBD$''. 
Applying Lemma~\ref{lem:variance_G}, we know that $\E[p_2(G)^2]\leq\beta^2 ={\(C_1\cdot(1+\log(\frac{1}{\alpha}))\cdot\log^2(2+\log(\frac{1}{\alpha}))\)^{2\ell} }$. Therefore, $\Var[\frac{1}{\beta}\cdot p_2(G)] \leq \frac{1}{\beta^2}\cdot\E[p_2(G)^2] \leq 1$ and thus satisfies the preconditions of \bmf. Then, if \bmf also returns ``$\TBD$'', we can show that $\Var[\frac{1}{\beta}\cdot p_2(T)] = O\big({(\ell+\log(\frac{1}{\alpha}))}^{2\ell} \cdot\log^2(2+\log(\frac{1}{\alpha}))\big)$ according to Theorem~\ref{thm:bmf}. Thus,
\begin{align*}
\Var[p_2(T)] &\leq \beta^2\cdot O(\ell+\log({1}/{\alpha}))^{2\ell}\log^2(2+\log({1}/{\alpha})) \\
&{ \leq O((\ell+\log({1}/{\alpha}))\log^2(2+\log({1}/{\alpha})))^{4\ell}. }
\end{align*} 
We then show by contradiction. Assume the above holds and $T$ is $\alpha$-good. Due to Theorem~\ref{thm:bmf}, we can show that 
\begin{align*}
\abs{\E[p_2(G)] - \E[p_2(T)]} 
&\leq \beta\cdot O(\ell+\log({1}/{\alpha}))^{\ell}\log(2+\log({1}/{\alpha})) \\
&{ \leq O((\ell+\log({1}/{\alpha}))\log^2(2+\log({1}/{\alpha})))^{2\ell}. }
\end{align*}
Additionally, since 
\begin{equation*}
\abs{\E[p_2(G)]} \leq \sqrt{\E[p_2^2(G)]} \leq \beta^2 = { O((\ell+\log({1}/{\alpha}))\log^2(2+\log({1}/{\alpha})))^{2\ell}, }
\end{equation*}
Therefore, by Cauchy-Schwarz inequality, we conclude that {$\abs{\E[p_2(T)]} \leq O((\ell+\log(\frac{1}{\alpha}))\log^2(2+\log(\frac{1}{\alpha})))^{2\ell}$}.
However, by construction, we have 
\begin{align*}
\abs{\E[p_2(T)]} &= \E\[\trace{\frac{(\tilde{\Sigma})_U}{\fronorm{(\tilde{\Sigma})_U}} \(P_{d,\ell}(T-\mu_T)P_{d,\ell}(T-\mu_T)^{\top}\)}\]\\
&= \trace{(\tilde{\Sigma})_U \tilde{\Sigma}}= \fronorm{(\tilde{\Sigma})_U} \stackrel{\zeta_4}{\geq} \lambdasparse  \\
&\geq C_1\cdot((\ell+ C_1\log({1}/{\alpha}))\log^2(2+\log({1}/{\alpha})))^{2\ell},\\
\end{align*}
where $\zeta_4$ is due to the condition in Step~\ref{step:MF-return-mean} of Algorithm~\ref{alg:mainSub-highdegree}. This is a contradiction. 

Hence, we conclude that $T$ cannot be $\alpha$-good and we remove it from the list. 
Moreover, if \bmf returns $\NO$ or a list $\{(T_i,\alpha_i)\}$, the guarantees follow from Theorem~\ref{thm:bmf}. The proof is complete.
\end{proof}

\begin{lemma}[Algorithm~\ref{alg:hmf}]\label{lem:hmf}
Consider Algorithm~\ref{alg:hmf} with input polynomial being $p_1$ or $p_2$ in view of Algorithm~\ref{alg:mainSub-highdegree}, and denote by $\ANS$ its output. With probability $1-\tau$, the following holds. If $\ANS = \NO$, then $T$ is not $\alpha$-good. If $\ANS = \{(T_i, \alpha_i)\}_{i=1}^m$ for some $m \leq 2$, then $T_i \subset T$ for all $i \in [m]$ and $\sum_{i=1}^{m} \frac{1}{\alpha_i^2} \leq \frac{1}{\alpha^2}$; if additionally $T$ is $\alpha$-good, then at least one $T_i$ is $\alpha_i$-good. 
\end{lemma}
\begin{proof}
Inside any subroutine of \bmf or \mmf called by \harmf, if $\ANS$ is assigned ``$\NO$'' or a list of pairs $\{(T_i,\alpha_i)\}$, the guarantees are ensured by Lemma~\ref{lem:variance_G}, Theorem~\ref{thm:bmf} and Lemma~\ref{lem:multilinear}. It remains to show the correctness of the algorithm returning ``$\NO$'' at Step~\ref{step:HMF-certified} when \bmf returns $\TBD$, which is implied by Lemma~\ref{lem:quadratic_filter}.
\end{proof}


\subsection{Proof of Theorem~\ref{thm:mainsub}}

\begin{proof}
The theorem follows from Lemma~\ref{lem:returned_vector}, Lemma~\ref{lem:linear_filter}, Lemma~\ref{lem:quadratic_filter}, Theorem~\ref{thm:bmf} and Lemma~\ref{lem:hmf}.
\end{proof}

\section{Proof of Theorem~\ref{thm:main_meanEst}}

Theorem~\ref{thm:main_meanEst} directly follows from the guarantees of our initial clustering step (Lemma~\ref{lem:cluster}), the main subroutine (Theorem~\ref{thm:mainsub}), and the black-box list reduction algorithm (Proposition~\ref{prop:reduce}).



\begin{proof}[Proof of Theorem~\ref{thm:main_meanEst}]

Consider Algorithm~\ref{alg:listdecodable}. By Lemma~\ref{lem:cluster}, $T$ will be divided into at most $\frac{1}{2\alpha}$ number of subsets, at least one of which is $\frac{\alpha}{2}$-good. Algorithm~\ref{alg:listdecodable} then maintains a list $\calL$ of pairs $\{(T_i,\alpha_i)\}$ on which Algorithm~\ref{alg:mainSub-highdegree} is called repetitively until the list becomes empty. Theorem~\ref{thm:mainsub} implies that when Algorithm~\ref{alg:mainSub-highdegree} is called on some $T_i\in\calL$ which is $\alpha_i$-good, if a list of pairs $\{(T_{j},\alpha_{j})\}$ is returned, then at least one of $\{T_j\}$ is $\alpha_{j}$-good ($\alpha_j>\alpha_i$). This ensures that there always exists an $\frac{\alpha}{2}$-good subset $T_i$ in list $\calL$, except that a leaf node has been created for this branch and the empirical mean of an $\frac{\alpha}{2}$-good data set is returnd.
We then argue that Algorithm~\ref{alg:listdecodable} eventually returns an estimated mean at the branch that includes only $\frac{\alpha}{2}$-good subsets. Since the subsets are $\frac{\alpha}{2}$-good, $\ANS$ never equals to $\NO$. In addition, the branch will not create child nodes forever: note that the true multifiltering step is in \bmf, and both Step~\ref{step:BMF-onecluster-filter} and \ref{step:BMF-twocluster-threshold} reduce the subset size $\abs{T_i}$ by at least $1$; since $\alpha_i$ is non-decreasing, the algorithm cannot remove only inliers; by Definition~\ref{def:good-set}, $\abs{\Sgood\cap T_i}\geq (1-\alpha_i/6)\abs{\Sgood}\geq\frac12\abs{\Sgood}$. Therefore, the algorithm must return an estimated mean when there is no outliers to filter.

We then bound the list size of the returned list of estimated means. Since during the process of multifiltering, $\sum_i \alpha_i^{-2}$ is non-increasing, we have that $\sum_{i=1}^{\abs{\calL}}\alpha_i^{-2}\leq\frac{1}{2\alpha}\cdot\alpha^{-2}$ at any point of Algorithm~\ref{alg:listdecodable}. In addition, $\alpha_i\leq1,\forall i$, meaning that the list size will never be larger than $O(\alpha^{-3})$. So does the size of $M$. Then, by applying \listred on $M$ with $\abs{M}\leq O(\alpha^{-3})$, the list size can be reduced to $O(\alpha^{-1})$ in view of Proposition~\ref{prop:reduce}.

Finally, note that \cluster runs in time $O\big(\poly(\abs{T},d)\big)$, \aefilter runs in time $O\big(\poly(\abs{T}, d^{\ell})\big)$ in view of Theorem~\ref{thm:mainsub}, and \listred runs in time $O\big(\poly(\abs{T},d)\big)$. Moreover, there are at most $O({\abs{T}}/{\alpha^3})$ number of calls to \aefilter, and only one call to \cluster and one call to \listred, we conclude that the time complexity of Algorithm~\ref{alg:listdecodable} is $O\big(\poly\big(\abs{T},d^{\ell},\frac{1}{\alpha}\big)\big)$.
\end{proof}


\section{Omitted Algorithms}
\label{sec:app:omit-alg}

In the following, we present the omitted algorithms. In particular, \mmf (Algorithms~\ref{alg:multilinear}) is an important component of \harmf, for which we tailor the algorithms in \cite{dia2018list} to our sparse setting. \mmf will further invoke \deghom (Algorithm~\ref{alg:degree2}). Algorithm~\ref{alg:reduce}, due to \cite{dia2018list}, is the black-box list reduction approach that was invoked in Algorithm~\ref{alg:listdecodable}.

\subsection{\mmf}

We introduce useful facts about multilinear polynomial here. For $d, l \in \mathbb{N}$, a polynomial $p(x_1,\dots,x_l):\R^{dl}\rightarrow\R$, where $x_i\in\R^d$, is called multilinear if it is linear in each of its $l$ arguments, i.e. if holds that $p(a\cdot x_1+b\cdot x_1',x_2,\dots,x_l) = a\cdot p(x_1,x_2,\dots,x_l) + b\cdot p(x_1,\dots,x_l)$, for all $a,b \in\R$ and $x_i,x_i'\in\R^d$, and similarly for all the other arguments. Moreover, a polynomial $p$ is called symmetric if $p(x_1, \dots, x_l) = p(x_{\pi(1)},\dots,x_{\pi(l)})$ for any permutation $\pi:[l]\rightarrow[l]$. Any degree-$l$ multilinear polynomial $p:\R^{dl}\rightarrow\R$ can be expressed as $A(x_1,\dots,x_l)$ for an order-$l$ tensor $A$ over $\R^d$. Moreover, $A$ is symmetric if $p$ is symmetric.

\begin{algorithm}[htb!]
\caption{\mmf}
\label{alg:multilinear}
\begin{algorithmic}[1]
	
	\REQUIRE A multiset of samples $T\subset \R^d$, parameter $\alpha\in(0,1/2]$, failure probability $\tau \in (0, 1)$, a degree-$l$ multilinear polynomial $V(x_1,\dots,x_l)$ over $\R^{dl}$ with $\twonorm{V}\leq1$, where $V$ is {the outer product of $l$ number of $\psi$-sparse vectors}.

	\STATE If {$l=1$}, run \bmf on $V(x-\mu_T)$, and {\bf return} its output.

	\STATE Compute the quadratic polynomial $q(x)=\twonorm{V(x-\mu_T)}^2$, where $x\in\R^d$ and $Vx$ is an order-$(l-1)$ tensor with $(Vx)_{i_2,\dots,i_l} = \sum_{i_1}x_{i_1}V_{i_1,\dots,i_l} $.

	\STATE Run \deghom on $q(x)$. If it returns $\NO$ or a list $\{T_i,\alpha_i\}$, then {\bf return} the same result.
	
	\STATE Sample a set $\Phi$ of $m=200\cdot\alpha^{-1}\log(4/\tau)$ instances uniformly at random from $T$. 
	
	\STATE $\forall x\in\Phi$, let $V_x = \frac{1}{\sqrt{q(x)}}\cdot V(x-\mu_T)$. $\ANS \gets$ \mmf on $(T,Vx,l-1,\alpha,\tau/2)$. If it returns $\NO$ or a list $\{(T_i,\alpha_i)\}$, then {\bf return} the same result. 
	
	\STATE Otherwise, {\bf return} $\TBD$.

\end{algorithmic}
\end{algorithm}

\begin{algorithm}[t]
\caption{$\deghom(T, \alpha, \tau, A)$}
\label{alg:degree2}
\begin{algorithmic}[1]
	\REQUIRE A multiset of samples $T\subset \R^d$, parameter $\alpha\in(0,1/2]$, failure probability $\tau \in (0, 1)$, homogeneous polynomial $x^{\top}Ax$, where $A$ is a $d\times d$ matrix with $\nuclearnorm{A}\leq1$. 
	
	
	\STATE  Compute the $k^2$ largest eigenvalues $\lambda_i$ and eigenvectors $v_i$ of $A$.
	
	\FOR{$i = 1, \dots, k^2$}
	
	\STATE $\ANS_i \gets \bmf(T, \alpha, \tau, p)$ with $p(x) = v_i \cdot x$.
	
	\STATE {\bfseries if} $\ANS_i$ is a list of $\{T_i,\alpha_i\}$ or $\ANS = \NO$ {\bfseries then return} $\ANS_i$.
	
	\ENDFOR
	
	\STATE {\bfseries if} all $\ANS_i = \TBD$ {\bfseries then return} $\TBD$.
	
\end{algorithmic}
\end{algorithm}

The \mmf works in the following way: Given degree-$l$ multilinear polynomial $V(x_1-\mu_T,\dots,x_l-\mu_T)$, where $V$ is an order-$l$ symmetric tensor and $x_i$'s are $l$ number of independent variables. The goal is to show that the polynomial has small absolute expectation over $l$ number of i.i.d. draws from $G\sim\N(\mu,\I_d)$ if the algorithm does not filter any samples and returns ``$\TBD$''; and otherwise, the algorithm multifilters correctly. Since all subroutine of \mmf multifilter the data set by calling \bmf on linear polynomials, we know that if it returns ``$\NO$'' or a list of pairs $\{(T_i,\alpha_i)\}$, the correctness is guaranteed by Theorem~\ref{thm:bmf}. It remains to bound the expected value of the multilinear polynomial when all subroutines return ``$\TBD$'' and $T$ is $\alpha$-good.

The idea is to sub-sample a large enough sample $\Phi$ from $T$. If $T$ is $\alpha$-good, then with sufficiently high probability, $\exists x\in\Phi$ that is from $G$. By recursively doing this, with sufficiently high probability, we construct a multilinear polynomial $V(G_1-\mu_T,G_2-\mu_T,\dots,G_{l}-\mu_T)$, the expectation of which is what we concerned about. The upper bound is then shown by induction. When the polynomial is linear, $\abs{\E[V_{x^{l-1}}(G-\mu_T)]}\leq O(\sqrt{1+\log({1}/{\alpha})}\cdot\log(2+\log({1}/{\alpha}))$. Here, we use $V_{x^i}$ to denote taking $i$ times of inner product between tensor $V$ and a vector $x$. Note that $x$ can be different in each time of the inner product. Then, without loss of generality, assume that for order-$(l-1)$ tensor $V_{x}$, $\abs{\E[V_{x}(G_1-\mu_T,\dots,G_{l-1}-\mu_T)]}\leq f(l-1,\alpha)$, we can show that $\abs{\E[V(G_1-\mu_T,\dots,G_{l}-\mu_T)]}\leq f(l,\alpha)$, provided that $\Sgood$ is sufficiently epresentative with respect to $G$ on any linear polynomial $V_{x^{l-1}}(x-\mu_T)$ and any quadratic polynomial $q(x) = \twonorm{V_{x^{i}}(x-\mu_T)}^2,\forall i\in[l]$. In this analysis, the only difference between our setting and that of~\cite{dia2018list} is the definition of representative set $\Sgood$ (Definition~\ref{def:representative-set}). Fortunately, since all polynomials in our algorithm apply to at most $\psi=2\ell k^{2\ell}$ coordinates, the linear polynomials must be in $\mathbb{P}(\R^d,1,2\ell k^{2\ell},2\ell k^{2\ell})$, and the quadratic polynomials must be in $\mathbb{P}(\R^d,2,4\ell^2 k^{4\ell},2\ell k^{2\ell})$. Therefore, our definition of representative set sufficies. The proof follows the same pipeline as that of Lemma~3.27 in~\cite{dia2018list}. As a result, it can be shown that $f(l,\alpha) = f(l-1,\alpha)\cdot O(\sqrt{1+\log({1}/{\alpha})}\cdot\log(2+\log({1}/{\alpha}))$, which renders $\abs{\E[V(G_1-\mu_T,\dots,G_{l}-\mu_T)]}\leq O\big(({1+\log({1}/{\alpha})})^{l/2}\cdot\log^{l}(2+\log({1}/{\alpha})\big)$.

\begin{definition}[Multifilter condition]\label{def:condition}
We say that a list of pairs $\{(T_i,\alpha_i)\}$, where $T_i\subset T$ and $\alpha_i\in(0,1)$, satisfies the multifilter condition for $(T,\alpha)$ if the following hold:
\begin{enumerate}
\item $\sum_i \frac{1}{\alpha_i^2}\leq \frac{1}{\alpha^2}$, and
\item If $T$ is $\alpha$-good, then at least one $T_i$ is $\alpha_i$-good.
\end{enumerate}
\end{definition}

\begin{lemma}[\mmf, Lemma~3.27 of~\cite{dia2018list}]\label{lem:multilinear}
Given $\alpha\in(0,\frac12]$ and $\tau\in(0,1)$, let $T$ be the input sample set, and a degree-$l$ multilinear polynomial $V(x_1,\dots,x_l)$ over $\R^{dl}$ with $\twonorm{V}=1$. Algorithm~\ref{alg:multilinear} returns one of the following with guarantees:
(1) $\TBD$, and we have that, if $T$ is $\alpha$-good, then with probability $1-\tau$, $\abs{\E[V(G_1-\mu_T, \dots, G_l-\mu_T)]} = O\big((1+\log(\frac{1}{\alpha})\log^2(2+\log(\frac{1}{\alpha})))\big)^{l/2}$, where $G_i$ are independent copies of $G$.
(2) $\NO$, then $T$ is not  $\alpha$-good. 
(3) A list of pairs $\{(T_i,\alpha_i)\}$, $T_i\subset T$,  satisfying the multifilter condition for $(T,\alpha)$.
\end{lemma}

\subsection{\listred}

\begin{algorithm}[htb!]
\caption{$\listred(T, \alpha, \ell, M)$}
\label{alg:reduce}
\begin{algorithmic}[1]
\REQUIRE A multiset of samples $T\subset\R^d$, parameter $\alpha \in (0, 1/2]$, degree $\ell \geq 1$, a list $M \subset \R^d$.

\STATE $\beta \assign C_4\cdot\alpha^{-\frac{1}{2\ell}}\sqrt{\ell}(\ell+\log\frac{1}{\alpha})$, $\delta \assign \frac{1}{C_5\log\frac{1}{\alpha}}$, $t\assign\sqrt{\log(C_5\log\frac{1}{\alpha})}$, $n\assign\abs{M}$.

\STATE For all $\mu_i,\mu_j \in M$, let $v_{ij}$ denote the unit vector in the $\mu_i-\mu_j$ direction.

\STATE  Let $T_i = \cap_{j\neq i}\{x\in T:\abs{v_{ij}\cdot(x-\mu_i)}<\beta+t \}$.

\STATE  $M'\assign\emptyset$.

\STATE $\forall i\in[n]$, if $\abs{T_i}\geq\alpha(1-\delta n)\abs{T}$, and $\nexists\mu_j\in M'$ such that $\twonorm{\mu_i-\mu_j}<2(\beta+t)$, then $M'\assign M'\cup\mu_i$.

\RETURN  $M'$.
\end{algorithmic}
\end{algorithm}

\begin{proposition}[\listred, Proposition~B.1 of~\cite{dia2018list}]\label{prop:reduce}
Fix $\alpha,\beta,\delta, t>0$ and let $\mu^*\in\R^d$ be finite, and let $S\subseteq T$ be so that (i) $\abs{S}/\abs{T}\geq\alpha$, and (ii) for all unit vectors $v\in\R^d$, we have $\Pr[v\cdot(S-\mu^*)>t]<\delta$. Then, given $M=\{\mu_1,\dots,\mu_n\}\subset\R^d$ so that $\delta n=o(1)$ and there is some $i$ so that $\twonorm{\mu_i-\mu^*}\leq\beta$ for some $\mu_i\in M$, Algorithm~\ref{alg:reduce} outputs $M'\subseteq M$ so that $\abs{M'}\leq\frac{1}{\alpha}(1+O(\delta n))$ and $\twonorm{\mu'-\mu^*}\leq 3(\beta+t)$ for some $\mu'\in M'$.
\end{proposition}

\begin{remark}
Under the setting of Algorithm~\ref{alg:reduce}, we have that $n=O(\alpha^{-3})$. This combined with the parameter settings in \listred shows that the size of $M'$ is $O(1/\alpha)$ and there is at least one $\mu_i \in M'$ that has comparable error guarantee to those in $M$.
\end{remark}

\section{Useful Lemmas}

\begin{lemma}[Lemma~3.2 of~\cite{cheng2021outlier}]\label{lem:sparse_vectors}
Fix two vectors $x,y$ with $\zeronorm{x}\leq k$ and $\twonorm{ \hard_k(x-y)}\leq\delta$. We have that $\twonorm{x-\hard_k(y)}\leq \sqrt{5}\delta$.
\end{lemma}

\begin{lemma}[degree-$l$ Chernoff bound, Fact~2.8 of~\cite{dia2018list}]\label{lem:chernoff}
Let $G\sim N(\mu,\I_d)$, $\mu\in\R^d$. Let $p:\R^d\rightarrow\R$ be a  degree-$l$ polynomial. For any $t>0$, we have that $\Pr\big[\abs{p(G)-\E[p(G)]} \geq t\cdot\sqrt{\Var[p(G)]}\big] \leq \exp\big(-\Omega(t^{2/l})\big)$.
\end{lemma}

\begin{lemma}[Harmonic and multilinear polynomials, Lemma~3.24 of~\cite{dia2018list}]\label{lem:harmonic_to_multilinear}
Let $X,X_{(1)},\dots,X_{(\ell)}$ be i.i.d random variables distributed as $\N(\mu,\I)$ for some $\mu\in\R^d$. Then, for any symmetric matrix $A$, we have
\begin{equation*}
\sqrt{\ell!}\cdot\E[h_A(X)] = \homogeneous_A(\mu) = \E[A(X_{(1)},\dots,X_{(\ell)})],
\end{equation*}
and
\begin{equation*}
\E[h_A(X)^2] = \sum_{\ell'=0}^{\ell}\(\binom{\ell}{\ell-\ell'}/\ell'!\) \cdot \homogeneous_{B^{(\ell')}}(\mu)
\end{equation*}
where $B^{(\ell')}$ is the order-$2\ell'$ tensor with 
\begin{equation*}
B^{(\ell')}_{i_1,\dots,i_{\ell'},j_1,\dots,j_{\ell'}} = \sum_{k_{\ell'+1},\dots,k_{\ell}} A_{i_1\dots,i_{\ell'},k_{\ell'+1},\dots,k_\ell}A_{j_1\dots,j_{\ell'},k_{\ell'+1},\dots,k_\ell}.
\end{equation*}
\end{lemma}

%% file: list-arxiv.bbl
\newcommand{\etalchar}[1]{$^{#1}$}
\begin{thebibliography}{DKK{\etalchar{+}}21b}

\bibitem[ABHM17]{awasthi2017efficient}
Pranjal Awasthi, Avrim Blum, Nika Haghtalab, and Yishay Mansour.
\newblock Efficient {PAC} learning from the crowd.
\newblock In {\em Proceedings of the 30th Annual Conference on Learning
  Theory}, pages 127--150, 2017.

\bibitem[ABHZ16]{awasthi2016learning}
Pranjal Awasthi, Maria{-}Florina Balcan, Nika Haghtalab, and Hongyang Zhang.
\newblock Learning and 1-bit compressed sensing under asymmetric noise.
\newblock In {\em Proceedings of the 29th Annual Conference on Learning
  Theory}, pages 152--192, 2016.

\bibitem[ABL17]{awasthi2017power}
Pranjal Awasthi, Maria{-}Florina Balcan, and Philip~M. Long.
\newblock The power of localization for efficiently learning linear separators
  with noise.
\newblock {\em Journal of the {ACM}}, 63(6):50:1--50:27, 2017.

\bibitem[AL87]{angluin1987learning}
Dana Angluin and Philip~D. Laird.
\newblock Learning from noisy examples.
\newblock {\em Machine Learning}, 2(4):343--370, 1987.

\bibitem[BDLS17]{balakrishnan2017computation}
Sivaraman Balakrishnan, Simon~S. Du, Jerry Li, and Aarti Singh.
\newblock Computationally efficient robust sparse estimation in high
  dimensions.
\newblock In {\em Proceedings of the 30th Annual Conference on Learning
  Theory}, pages 169--212, 2017.

\bibitem[BHL95]{blum1995learning}
Avrim Blum, Lisa Hellerstein, and Nick Littlestone.
\newblock Learning in the presence of finitely or infinitely many irrelevant
  attributes.
\newblock {\em Journal of Computer and System Sciences}, 50(1):32--40, 1995.

\bibitem[BJK15]{bhatia2015robust}
Kush Bhatia, Prateek Jain, and Purushottam Kar.
\newblock Robust regression via hard thresholding.
\newblock In {\em {NIPS}}, pages 721--729, 2015.

\bibitem[BK21]{bakshi2021list}
Ainesh Bakshi and Pravesh~K. Kothari.
\newblock List-decodable subspace recovery: Dimension independent error in
  polynomial time.
\newblock In {\em Proceedings of the 2021 {ACM-SIAM} Symposium on Discrete
  Algorithms}, pages 1279--1297, 2021.

\bibitem[CDK{\etalchar{+}}21]{cheng2021outlier}
Yu~Cheng, Ilias Diakonikolas, Daniel~M. Kane, Rong Ge, Shivam Gupta, and Mahdi
  Soltanolkotabi.
\newblock Outlier-robust sparse estimation via non-convex optimization.
\newblock {\em CoRR}, abs/2109.11515, 2021.

\bibitem[CDS98]{chen1998atomic}
Scott~Shaobing Chen, David~L. Donoho, and Michael~A. Saunders.
\newblock Atomic decomposition by basis pursuit.
\newblock {\em {SIAM} Journal on Scientific Computing}, 20(1):33--61, 1998.

\bibitem[CMY20]{cmy2020list}
Yeshwanth Cherapanamjeri, Sidhanth Mohanty, and Morris Yau.
\newblock List decodable mean estimation in nearly linear time.
\newblock In {\em 61st {IEEE} Annual Symposium on Foundations of Computer
  Science}, pages 141--148. {IEEE}, 2020.

\bibitem[CSV17]{charika2017learning}
Moses Charikar, Jacob Steinhardt, and Gregory Valiant.
\newblock Learning from untrusted data.
\newblock In {\em Proceedings of the 49th Annual {ACM} {SIGACT} Symposium on
  Theory of Computing}, pages 47--60, 2017.

\bibitem[CT05]{candes2005decoding}
Emmanuel~J. Cand{\`{e}}s and Terence Tao.
\newblock Decoding by linear programming.
\newblock {\em {IEEE} Transactions on Information Theory}, 51(12):4203--4215,
  2005.

\bibitem[DKK{\etalchar{+}}16]{diakonikolas2016robust}
Ilias Diakonikolas, Gautam Kamath, Daniel~M. Kane, Jerry Li, Ankur Moitra, and
  Alistair Stewart.
\newblock Robust estimators in high dimensions without the computational
  intractability.
\newblock In {\em Proceedings of the 57th Annual {IEEE} Symposium on
  Foundations of Computer Science}, pages 655--664, 2016.

\bibitem[DKK{\etalchar{+}}17]{dia2017being}
Ilias Diakonikolas, Gautam Kamath, Daniel~M. Kane, Jerry Li, Ankur Moitra, and
  Alistair Stewart.
\newblock Being robust (in high dimensions) can be practical.
\newblock In {\em Proceedings of the 34th International Conference on Machine
  Learning}, volume~70 of {\em Proceedings of Machine Learning Research}, pages
  999--1008. {PMLR}, 2017.

\bibitem[DKK{\etalchar{+}}19]{dia2019outlier}
Ilias Diakonikolas, Daniel Kane, Sushrut Karmalkar, Eric Price, and Alistair
  Stewart.
\newblock Outlier-robust high-dimensional sparse estimation via iterative
  filtering.
\newblock In {\em NeurIPS}, pages 10688--10699, 2019.

\bibitem[DKK20a]{dia2020list}
Ilias Diakonikolas, Daniel Kane, and Daniel Kongsgaard.
\newblock List-decodable mean estimation via iterative multi-filtering.
\newblock In {\em Proceedings of the 34th Annual Conference on Neural
  Information Processing Systems}, 2020.

\bibitem[DKK{\etalchar{+}}20b]{diakonikolas2020polynomial}
Ilias Diakonikolas, Daniel~M. Kane, Vasilis Kontonis, Christos Tzamos, and
  Nikos Zarifis.
\newblock A polynomial time algorithm for learning halfspaces with {Tsybakov}
  noise.
\newblock {\em CoRR}, abs/2010.01705, 2020.

\bibitem[DKK{\etalchar{+}}21a]{dia2021list}
Ilias Diakonikolas, Daniel Kane, Daniel Kongsgaard, Jerry Li, and Kevin Tian.
\newblock List-decodable mean estimation in nearly-pca time.
\newblock In {\em Proceedings of the 35th Annual Conference on Neural
  Information Processing Systems}, pages 10195--10208, 2021.

\bibitem[DKK{\etalchar{+}}21b]{dia2021cluster}
Ilias Diakonikolas, Daniel~M. Kane, Daniel Kongsgaard, Jerry Li, and Kevin
  Tian.
\newblock Clustering mixture models in almost-linear time via list-decodable
  mean estimation.
\newblock {\em CoRR}, abs/2106.08537, 2021.

\bibitem[DKK{\etalchar{+}}22]{diaLDsparse2022}
Ilias Diakonikolas, Daniel~M. Kane, Sushrut Karmalkar, Ankit Pensia, and
  Thanasis Pittas.
\newblock List-decodable sparse mean estimation via difference-of-pairs
  filtering.
\newblock {\em CoRR}, abs/2206.05245, 2022.

\bibitem[DKS17]{diakonikolas2017statistical}
Ilias Diakonikolas, Daniel~M. Kane, and Alistair Stewart.
\newblock Statistical query lower bounds for robust estimation of
  high-dimensional gaussians and gaussian mixtures.
\newblock In {\em Proceedings of the 58th {IEEE} Annual Symposium on
  Foundations of Computer Science}, pages 73--84, 2017.

\bibitem[DKS18a]{diakonikolas2018learning}
Ilias Diakonikolas, Daniel~M. Kane, and Alistair Stewart.
\newblock Learning geometric concepts with nasty noise.
\newblock In {\em Proceedings of the 50th Annual {ACM} Symposium on Theory of
  Computing}, pages 1061--1073, 2018.

\bibitem[DKS18b]{dia2018list}
Ilias Diakonikolas, Daniel~M. Kane, and Alistair Stewart.
\newblock List-decodable robust mean estimation and learning mixtures of
  spherical gaussians.
\newblock In {\em Proceedings of the 50th Annual {ACM} {SIGACT} Symposium on
  Theory of Computing}, pages 1047--1060. {ACM}, 2018.

\bibitem[DKTZ20]{diakonikolas2020learning}
Ilias Diakonikolas, Vasilis Kontonis, Christos Tzamos, and Nikos Zarifis.
\newblock Learning halfspaces with {Massart} noise under structured
  distributions.
\newblock In {\em Proceedings of the 33rd Annual Conference on Learning
  Theory}, pages 1486--1513, 2020.

\bibitem[Don06]{donoho2006compressed}
David~L. Donoho.
\newblock Compressed sensing.
\newblock {\em {IEEE} Transactions on Information Theory}, 52(4):1289--1306,
  2006.

\bibitem[Hau92]{haussler1992decision}
David Haussler.
\newblock Decision theoretic generalizations of the {PAC} model for neural net
  and other learning applications.
\newblock {\em Information and Computation}, 100(1):78--150, 1992.

\bibitem[HLZ20]{hopkins2020robust}
Samuel~B. Hopkins, Jerry Li, and Fred Zhang.
\newblock Robust and heavy-tailed mean estimation made simple, via regret
  minimization.
\newblock {\em CoRR}, abs/2007.15839, 2020.

\bibitem[Hub64]{huber1964robust}
Peter~J. Huber.
\newblock {Robust Estimation of a Location Parameter}.
\newblock {\em The Annals of Mathematical Statistics}, 35(1):73 -- 101, 1964.

\bibitem[KKK19]{karmalkar2019list}
Sushrut Karmalkar, Adam~R. Klivans, and Pravesh Kothari.
\newblock List-decodable linear regression.
\newblock In {\em Proceedings of the 33rd Annual Conference on Neural
  Information Processing Systems}, pages 7423--7432, 2019.

\bibitem[KKM18]{klivans2018efficient}
Adam~R. Klivans, Pravesh~K. Kothari, and Raghu Meka.
\newblock Efficient algorithms for outlier-robust regression.
\newblock In {\em Proceedings of the 31st Annual Conference on Learning
  Theory}, volume~75 of {\em Proceedings of Machine Learning Research}, pages
  1420--1430. {PMLR}, 2018.

\bibitem[KKMS05]{kalai2005agnostic}
Adam~Tauman Kalai, Adam~R. Klivans, Yishay Mansour, and Rocco~A. Servedio.
\newblock Agnostically learning halfspaces.
\newblock In {\em Proceedings of the 46th Annual {IEEE} Symposium on
  Foundations of Computer Science}, pages 11--20, 2005.

\bibitem[KL88]{kearns1988learning}
Michael~J. Kearns and Ming Li.
\newblock Learning in the presence of malicious errors.
\newblock In {\em Proceedings of the 20th Annual {ACM} Symposium on Theory of
  Computing}, pages 267--280, 1988.

\bibitem[KSS92]{kearns1992efficient}
Michael~J. Kearns, Robert~E. Schapire, and Linda Sellie.
\newblock Toward efficient agnostic learning.
\newblock In {\em Proceedings of the Fifth Annual {ACM} Conference on
  Computational Learning Theory}, pages 341--352, 1992.

\bibitem[KSS18]{kothari2018robust}
Pravesh~K. Kothari, Jacob Steinhardt, and David Steurer.
\newblock Robust moment estimation and improved clustering via sum of squares.
\newblock In {\em Proceedings of the 50th Annual {ACM} {SIGACT} Symposium on
  Theory of Computing}, pages 1035--1046, 2018.

\bibitem[Lit87]{littlestone1987learning}
Nick Littlestone.
\newblock Learning quickly when irrelevant attributes abound: {A} new
  linear-threshold algorithm.
\newblock In {\em Proceedings of the 28th Annual {IEEE} Symposium on
  Foundations of Computer Science}, pages 68--77, 1987.

\bibitem[LRV16]{lai2016agnostic}
Kevin~A. Lai, Anup~B. Rao, and Santosh~S. Vempala.
\newblock Agnostic estimation of mean and covariance.
\newblock In {\em Proceedings of the 57th Annual {IEEE} Symposium on
  Foundations of Computer Science}, pages 665--674, 2016.

\bibitem[LSLC20]{liu2020high}
Liu Liu, Yanyao Shen, Tianyang Li, and Constantine Caramanis.
\newblock High dimensional robust sparse regression.
\newblock In {\em The 23rd International Conference on Artificial Intelligence
  and Statistics,}, volume 108 of {\em Proceedings of Machine Learning
  Research}, pages 411--421. {PMLR}, 2020.

\bibitem[LT91]{ledoux1991probability}
Michel Ledoux and Michel Talagrand.
\newblock {\em Probability in Banach Spaces: Isoperimetry and Processes}.
\newblock Springer-Verlag Berlin Heidelberg, 1991.

\bibitem[Ma13]{ma2013sparse}
Zongming Ma.
\newblock Sparse principal component analysis and iterative thresholding.
\newblock {\em The Annals of Statistics}, 41(2):772--801, 2013.

\bibitem[MN06]{massart2006risk}
Pascal Massart and {\'E}lodie N{\'e}d{\'e}lec.
\newblock Risk bounds for statistical learning.
\newblock {\em The Annals of Statistics}, pages 2326--2366, 2006.

\bibitem[MV18]{meister2018data}
Michela Meister and Gregory Valiant.
\newblock A data prism: {S}emi-verified learning in the small-alpha regime.
\newblock In {\em Proceedings of the 31st Conference On Learning Theory}, pages
  1530--1546, 2018.

\bibitem[PV13]{plan2013robust}
Yaniv Plan and Roman Vershynin.
\newblock Robust 1-bit compressed sensing and sparse logistic regression: {A}
  convex programming approach.
\newblock {\em {IEEE} Transactions on Information Theory}, 59(1):482--494,
  2013.

\bibitem[RY20a]{raghavendra2020list}
Prasad Raghavendra and Morris Yau.
\newblock List decodable learning via sum of squares.
\newblock In {\em Proceedings of the 2020 {ACM-SIAM} Symposium on Discrete
  Algorithms}, pages 161--180, 2020.

\bibitem[RY20b]{raghavendra2020subspace}
Prasad Raghavendra and Morris Yau.
\newblock List decodable subspace recovery.
\newblock In {\em Proceedings of the 33rd Annual Conference on Learning
  Theory}, pages 3206--3226, 2020.

\bibitem[SCV18]{steinhardt2018resilience}
Jacob Steinhardt, Moses Charikar, and Gregory Valiant.
\newblock Resilience: {A} criterion for learning in the presence of arbitrary
  outliers.
\newblock In {\em Proceedings of the 9th Innovations in Theoretical Computer
  Science Conference}, pages 45:1--45:21, 2018.

\bibitem[She20]{shen2020one}
Jie Shen.
\newblock One-bit compressed sensing via one-shot hard thresholding.
\newblock In {\em Proceedings of the 36th Conference on Uncertainty in
  Artificial Intelligence}, pages 510--519, 2020.

\bibitem[She21]{shen2021sample}
Jie Shen.
\newblock Sample-optimal {PAC} learning of halfspaces with malicious noise.
\newblock In {\em Proceedings of the 38th International Conference on Machine
  Learning}, pages 9515--9524, 2021.

\bibitem[SL17a]{shen2017iteration}
Jie Shen and Ping Li.
\newblock On the iteration complexity of support recovery via hard thresholding
  pursuit.
\newblock In {\em Proceedings of the 34th International Conference on Machine
  Learning}, pages 3115--3124, 2017.

\bibitem[SL17b]{shen2017partial}
Jie Shen and Ping Li.
\newblock Partial hard thresholding: Towards a principled analysis of support
  recovery.
\newblock In {\em Proceedings of the 31st Annual Conference on Neural
  Information Processing Systems}, pages 3127--3137, 2017.

\bibitem[SL18]{shen2018tight}
Jie Shen and Ping Li.
\newblock A tight bound of hard thresholding.
\newblock {\em Journal of Machine Learning Research}, 18(208):1--42, 2018.

\bibitem[Slo88]{sloan1988types}
Robert~H. Sloan.
\newblock Types of noise in data for concept learning.
\newblock In {\em Proceedings of the First Annual Workshop on Computational
  Learning Theory}, pages 91--96, 1988.

\bibitem[STT12]{servedio2012attribute}
Rocco~A. Servedio, Li{-}Yang Tan, and Justin Thaler.
\newblock Attribute-efficient learning and weight-degree tradeoffs for
  polynomial threshold functions.
\newblock In {\em Proceedings of the 25th Annual Conference on Learning
  Theory}, pages 1--19, 2012.

\bibitem[SVC16]{steinhardt2016avoid}
Jacob Steinhardt, Gregory Valiant, and Moses Charikar.
\newblock Avoiding imposters and delinquents: Adversarial crowdsourcing and
  peer prediction.
\newblock In {\em Proceedings of the 30th Annual Conference on Neural
  Information Processing Systems}, pages 4439--4447, 2016.

\bibitem[SZ21]{shen2021attribute}
Jie Shen and Chicheng Zhang.
\newblock Attribute-efficient learning of halfspaces with malicious noise:
  Near-optimal label complexity and noise tolerance.
\newblock In {\em Proceedings of the 32nd International Conference on
  Algorithmic Learning Theory}, pages 1072--1113, 2021.

\bibitem[Tib96]{tibshirani1996regression}
Robert Tibshirani.
\newblock Regression shrinkage and selection via the {L}asso.
\newblock {\em Journal of the Royal Statistical Society: Series B
  (Methodological)}, 58(1):267--288, 1996.

\bibitem[Tuk60]{tukey1960survey}
John~W. Tukey.
\newblock A survey of sampling from contaminated distributions.
\newblock {\em Contributions to probability and statistics}, pages 448--485,
  1960.

\bibitem[Val85]{valiant1985learning}
Leslie~G. Valiant.
\newblock Learning disjunction of conjunctions.
\newblock In {\em Proceedings of the 9th International Joint Conference on
  Artificial Intelligence}, pages 560--566, 1985.

\bibitem[Wai19]{wainwright2019high}
Martin~J. Wainwright.
\newblock {\em High-dimensional statistics: {A} non-asymptotic viewpoint}.
\newblock Cambridge University Press, 2019.

\bibitem[WSL18]{wang2018provable}
Jing Wang, Jie Shen, and Ping Li.
\newblock Provable variable selection for streaming features.
\newblock In {\em Proceedings of the 35th International Conference on Machine
  Learning}, pages 5158--5166, 2018.

\bibitem[ZS21]{zeng2021semi}
Shiwei Zeng and Jie Shen.
\newblock Semi-verified learning from the crowd with pairwise comparisons.
\newblock {\em CoRR}, abs/2106.07080, 2021.

\bibitem[ZS22]{zeng2022crowd}
Shiwei Zeng and Jie Shen.
\newblock Efficient {PAC} learning from the crowd with pairwise comparisons.
\newblock In {\em Proceedings of the 39th International Conference on Machine
  Learning}, pages 25973--25993, 2022.

\bibitem[ZSA20]{zhang2020efficient}
Chicheng Zhang, Jie Shen, and Pranjal Awasthi.
\newblock Efficient active learning of sparse halfspaces with arbitrary bounded
  noise.
\newblock In {\em Proceedings of the 34th Annual Conference on Neural
  Information Processing Systems}, pages 7184--7197, 2020.

\end{thebibliography}
